\documentclass[lettersize,journal]{IEEEtran}
\usepackage{amsmath,amsfonts}
\usepackage{algorithmic}
\usepackage{array}
\usepackage[caption=false,font=normalsize,labelfont=sf,textfont=sf]{subfig}
\usepackage{textcomp}
\usepackage{stfloats}
\usepackage{url}
\usepackage{graphicx}
\usepackage{xcolor}

\usepackage{verbatim}

\def\BibTeX{{\rm B\kern-.05em{\sc i\kern-.025em b}\kern-.08em
    T\kern-.1667em\lower.7ex\hbox{E}\kern-.125emX}}
\usepackage{balance}

\usepackage{bm}
\usepackage{multirow}

\usepackage{CJK}

\begin{document}

\newcommand{\vd}{\dot{v}}
\newcommand{\yd}{\dot{y}}
\newcommand{\ydd}{\ddot{y}}
\newcommand{\xd}{\dot{x}}
\newcommand{\qd}{\dot{q}}
\newcommand{\qdd}{\ddot{q}}
\newcommand{\qddd}{\dddot{q}}
\renewcommand{\vd}{\dot{v}}
\newcommand{\w}{\mathbf{w}}
\newcommand{\A}{\mathbf{A}}
\renewcommand{\a}{\mathbf{a}}
\newcommand{\f}{\mathbf{f}}
\renewcommand{\P}{\mathbf{P}}
\IEEEoverridecommandlockouts
\IEEEpubid{\makebox[\columnwidth]{\copyright2022 IEEE \hfill} \hspace{\columnsep}\makebox[\columnwidth]{ }}



\title{Bootstrapping Concept Formation in Small Neural Networks}

\author{Minija Tamosiunaite,
        Tomas Kulvicius,
        and Florentin W\"org\"otter
\thanks{Acknowledgements: Supported by the European Community’s Horizon 2020 Programme grant number 899265, ADOPD.}
\thanks{M. Tamosiunaite, T. Kulvicius, and F. W\"org\"otter are with the Department for Computational Neuroscience, Third Physics Institute, University of G\"ottingen, 37073 G\"ottingen, Germany.}
\thanks{M. Tamosiunaite is also with the Faculty of Informatics, Vytautas Magnus University, Kaunas, Lithuania. email: minija.tamosiunaite@vdu.lt}
\thanks{T. Kulvicius is also with University Medical Center G\"ottingen, Child and Adolescent Psychiatry and Psychotherapy, 37075 G\"ottingen, Germany.}
}

\maketitle

\IEEEpubidadjcol

\begin{abstract}
The question how neural systems (of humans) can perform reasoning is still far from being solved. We posit that the process of forming Concepts is a fundamental step required for this. We argue that, first, Concepts are formed as closed representations, which are then consolidated by relating them to each other. Here we present a model system (agent) with a small neural network that uses realistic learning rules and receives only feedback from the environment in which the agent performs virtual actions. First, the actions of the agent are reflexive. In the process of learning, statistical regularities in the input lead to the formation of neuronal pools representing relations between the entities observed by the agent from its artificial world. This information then influences the behavior of the agent via feedback connections replacing the initial reflex by an action driven by these relational representations. We hypothesize that the neuronal pools representing relational information can be considered as primordial Concepts, which may in a similar way be present in some pre-linguistic animals, too. This system provides formal grounds for further discussions on what could be understood as a Concept and shows that associative learning is enough to develop concept-like structures.
\end{abstract}


\vspace{-7mm}
\section{Introduction}




\begin{CJK}{UTF8}{min}

How to clearly distinguish processes of reasoning and deliberation from reflexes in animals is unclear and it is also unknown at what level in the animal kingdom the former begins to emerge \cite{povinelli2020can}. Evolutionary origins as well as the organization of such processes in the human brain are largely unknown, too. Therefore, currently we have no means to equip robots and AI systems with such processes in a way that scales up to the level of human proficiency. In this paper, we are concerned with the question how deliberation could begin to emerge in a simple network from low-level neural principles. We posit that such processes must be fundamentally linked to the formation of Concepts in an agent upon which deliberation can commence. The goal of this paper is to show that small networks can perform the first small steps into this direction. Hence, we want to arrive at an algorithmically implementable theory that uses realistic neuronal operations working towards Concept formation.

The nature of Concepts is a subject of a long debate because it can be approached from different perspectives: e.g., language, psychology and neuroscience, but possibly it is studied most extensively in philosophy \cite{margolis2019concepts}. Concepts are frequently explained through examples given in natural language and are considered to be ``mental representations'' \cite{fodor1987psychosemantics,carey1992origin,pinker2003language}. However, such approaches are rarely mathematically precise nor do they suggest a procedure for algorithmic implementation. For our approach we get inspiration from the process of Peircean semiosis \cite{atkin2010peirce}, which is among the most constructive approaches in this field and leads us towards an algorithmic implementation.

\begin{figure}[ht]
\centering \includegraphics[width=5cm]{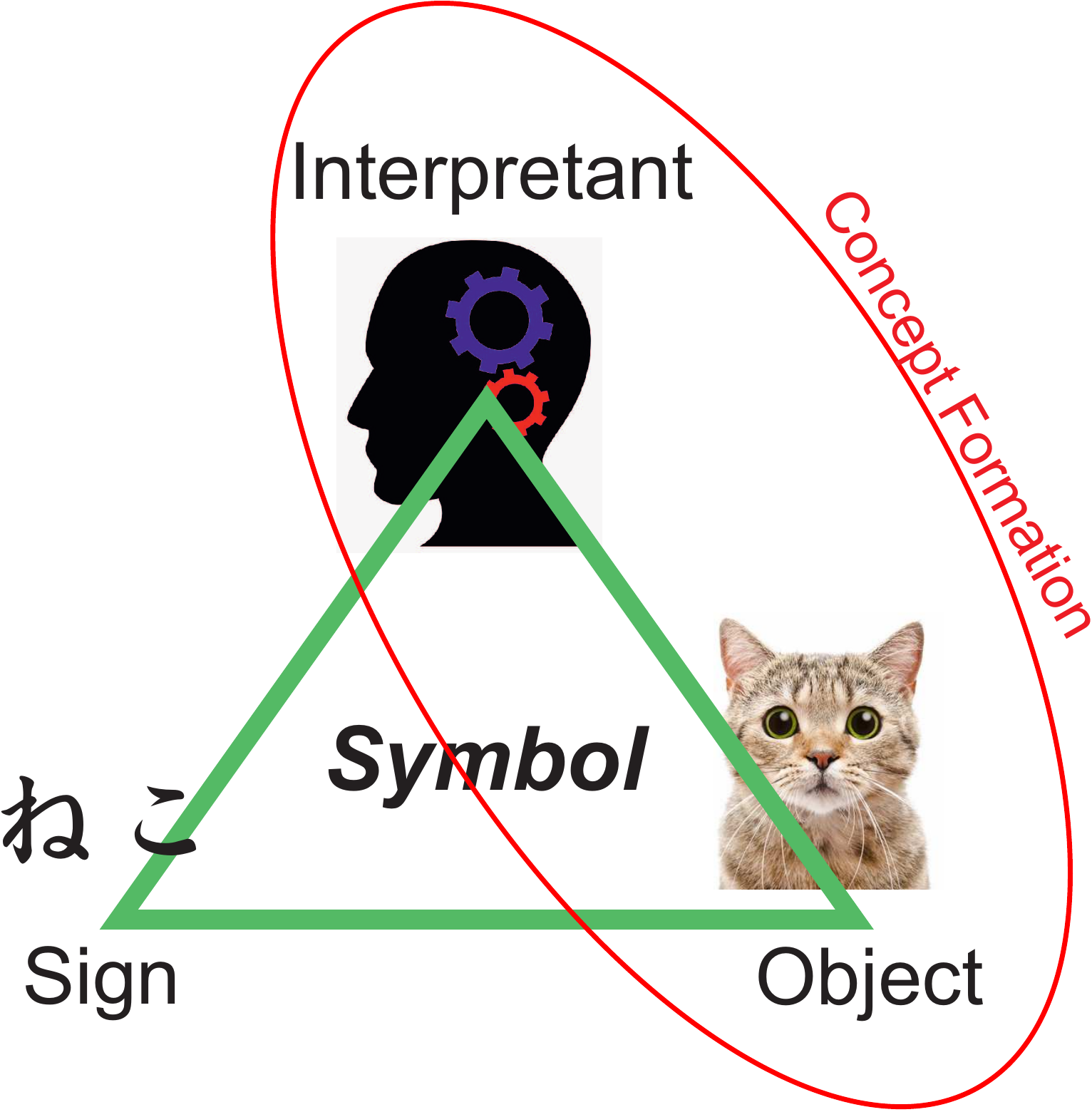}
\caption{Schematics of the process of semiosis.}
\label{semiosis}
\vspace*{-4mm}
\end{figure}

According to Peirce \cite{atkin2010peirce}, semiosis has three elements which are \textit{Sign}, \textit{Object}, and \textit{Interpretant}. Their triadic relationship is shown in Fig.~\ref{semiosis}. Sign represents a structure in the domain of perceivable signals, for example letters, utterances, etc.  Object is the entity to which the Sign refers, whereas Interpretant is the effect of a Sign on an interpreter (e.g., a person) who perceives the Sign. This triadic structure captures what we nowadays call a ``Symbol'' (e.g., in the context of development of communication between artificial agents). Semiotics is a complex multidisciplinary field and we cannot provide an in depth discussion here. Rather than that, we would like to ask about the neuronal processes underlying semiosis.


For our purposes one aspect in this discussion remains simple. The actual Sign will not matter. Call it cat, Katze, ねこ, chat, or xyz, as long as the interpreter (and possible communication partners) know which object concept stands behind these Signs all will be good.
Here we address  the hard problem to define the Interpretant. We ask the question, what is the Concept underlying the Object to which the Sign ``ねこ''
refers? Hence, we are concerned with the right side of this triangle  that creates the Concept.

The central problem addressed in this study is how to arrive at a primordial neuronal system that displays some of the processing steps needed for the formation of Concepts. Underlying this is the question, how advanced symbol-processing engines -- humans -- could have evolved from their ancestors. We would like to show here some simple neuronal processing steps that might belong to those that have kick-started the evolution towards advanced symbol processing and language.

We are here considering early (potential) stages in evolution. Thus, we are concerned with autopoietic systems \cite{varela1974autopoiesis}, because such agents were not yet supervised from ``someone else''. Thus, all learning therein should be based only on the statistical properties of the world. Such a system should be designed so that feedback comes only from the environment through the system's own behavior and own sensing capabilities \cite{Uexkull1992stroll}.

Hence, our system should carry within its activation patterns some characteristics that could be interpreted by an external observer as ``going towards representing a Concept'' and it should not just follow a chain of input-triggered, reflex-like actions to achieve a goal. Nonetheless, these ``Concept-like entities'' should arise solely from sensory inputs. As a consequence of this, behavior should then be triggered when certain activity patterns arise from the ``Concept assemblies'' and feed back to lower layers that are concerned with sensory-motor processing. The central problem that immediately shines up here is how to get from plain sensory signals to Concepts and back to motor signals. The heavy use of quotation marks above indicates that this primordial system will still remain quite far away from the symbol-handling capabilities found in apes, let alone humans.


To summarize, the contribution of this study is an artificial neural system that:

\begin{itemize}
\item  Forms Concept-like entities based on sensor (specifically, visual) inputs using associative learning.
\item Interprets the developed Concept-like entities by performing (virtual) actions appropriate for the given environmental condition.
\item Allows quantitative evaluation of performance through evaluation of action appropriateness.
\end{itemize}

From a machine learning perspective, our emphasis is on the problem of signal to symbol transformation based on associative learning. We argue that concept formation is an essential step in symbol formation. We hypothesize that concept formation can be bootstrapped  by associative learning, given input from the environment. We prove this hypothesis by showing that the system can behave appropriately based on concept-like entities, developed in an associative way. Admittedly, much of our conclusions will be speculative and matter of debate (see also Discussion), which we would like to stimulate with this article.


\section{State of the Art}

\underline{Symbol vs. signal.} Deliberation processes, as considered in this study, are most strongly related to language or classical AI,  both operating on discrete symbols \cite{sloman2014requirements}. However, this cannot be directly used in agents acting in the real world (e.g., robots), because both, sensors and motors, rely only on continuous (analog) signals. Thus, to use symbols with robots one needs a signal to symbol transformation. This problem had been addressed since the early days of AI in image- and other signal-processing applications in order to achieve a human-like analysis of a visual scene or of situations derived from other (non-visual) signal information \cite{minsky1961steps,winston1970learning,nii1982signal}.  Methods for covering the gap between signals and symbols many times are analyzed from the perspective of symbol grounding \cite{harnad1990symbol}; that is the process of  meaning acquisition of a symbol, where ``meaning'' itself is a complex concept \cite{harnad2007symbol}. In robotics, symbol grounding is important with respect to several aspects. First and most straightforward there is the aspect of communication with (or amongst) artificial agents where grounding of communicative symbols is needed for the agents to understand each other \cite{coradeschi2013short}. The second aspect is less straightforward and associated to the usage of layered architectures in cognitive robotic systems, where at the lowest layer sensori-motor signals are considered and at the highest layer symbolic planning is  performed \cite{gat1998three}. Thus, symbols first need to be derived from sensor signals to supply the planning procedures and then symbols coming from the planner need to be translated to (voltage-)signals for robotic motor-actions. All in all this comprises a pathway: signal-symbol-signal. Intriguingly,  the second aspect is  associated to ``speaking to yourself'', discussed in the philosophical literature \cite{dennett1993consciousness}. One can imagine a human mental plan to be a sequence of steps (tokens) kept in one's mind and communicated to yourself and (often) leading to (muscle-) actions. Thus, this links aspect two (layered architectures) back to aspect one (communication).

Symbol grounding in artificial agents is frequently addressed using multi-modal categorization in sensori-motor  space  and  statistical inference on top of that to associate the developed categories with words, i.e. symbols (e.g., told by a human) \cite{yu2004integration, nakamura2009grounding}. Such methods are applied in developmental robotics where robots can perform externalization of their perception and communication within the learned domain \cite{nakamura2009grounding,cangelosi2010grounding}. However until now only limited complexity is reached \cite{aly2018towards, el2021teaching}. For robots to perform more complicated tasks, instead of  grounding processes through self-experience, symbol anchoring  \cite{coradeschi2003introduction} is  supplied by the system's designer in a supervised  domain-based manner. Anchoring concentrates on creating and maintaining the relation between symbolic and sensory information, while grounding concentrates on the more advanced aspects of creating the ``meaning'' of symbols \cite{vogt2003anchoring} from first (sensori) principles. Thus, anchoring is more suited for robotic applications within well-defined domains where the designer has ``all'' knowledge and the robot never stumbles upon anything unexpected-by-the-designer (which it would, thus, not ``understand''). Hence, in such architectures, appropriate procedures are defined for world state estimation from sensing and planning where pre-programmed actions are then triggered \cite{wachter2018integrating,aein2019library}. Clearly, these approaches have only limited generalization abilities, mostly based on re-shuffling of the developed structures \cite{worgotter2015structural,  tamosiunaite2019cut} where also re-purposing of the structures using high level reasoning can be achieved to some degree \cite{tenorth2013knowrob,ramirez2017transferring}.

\underline{Connectionism and symbols.} An alternative way to arrive at situation-matched actions, as compared to classical AI, is by employing  end-to-end learning (sensor-to action), which is currently widely investigated \cite{finn2017deep,yang2020deep}. However, such approaches also do not reliably generalize to new situations and are hard to interpret, because sensor-to-action networks are difficult to comprehend. This none withstanding,  already Harnad \cite{harnad1990symbol} had advocated that connectionism - hence the use of networks - can serve for categorical representation using bottom-up approaches. Furthermore, we are a living proof of networks that use symbols: humans do reason symbolically and can manipulate symbols in an abstract way using their brain-networks. But what makes us different from animals that cannot do this? Penn et al. \cite{penn2008darwin} have suggested that it is the handling of relations between and on top of other relations that distinguishes the human from an animal's mind. This would correspond to a so-called \emph{physical symbol system} \cite{newell1980physical} which considers  abstract symbol manipulations of the type used in computers (and classical AI). However, using a neuronal implementation to target full-fledged abstract symbolic reasoning is still too ambitious for our currently existing artificial networks and it remains still too complex to attempt symbol grounding this way starting from sensory processing. Due to that, physical symbol systems are not much considered in recent studies of symbol emergence.

\underline{Concepts as units of thought.} Even if one does not target to explain the animal-to-human cognition leap \cite{penn2008darwin}, one still can  take inspiration from human thought processes and researchers debate how human thought is organized (e.g., does language come first or, alternatively, other aspects of reasoning, e.g., non-language based concepts?)  \cite{pinker2003language,tomasello2010origins}. The term ``Concept'' is here many times used to define structural units of thought. This term is mostly applied in philosophy, psychology, and linguistics, whereas introducing this term (and associated properties) into the realm of artificial systems is a relatively recent development. This, however, might be needed if one wants to organize an artificial system such that the processes therein resemble human thought.

In philosophy, Concepts are approached from several different perspectives, where the main trends are: mental images, abilities, or abstract objects \cite{margolis2019concepts}. Philosophers talk about mental representations arising through experience since centuries (e.g., see \cite{locke1847essay}).  In the mental image approach, Concepts are considered as building blocks of thought, where more complex representations are composed of simpler ones and where at the bottom reside primitive representations with perceptual character \cite{laurence1999concepts}. This, in principle, reflects the grounding processes addressed in computer science. Critique to the mental image approach arises due to difficulties in connecting abstract Concepts to perception. Alternative approaches span the domains of language and logic \cite{fodor1987psychosemantics,dummett1993seas,zalta2001fregean}, while assigning less importance to perception. We cannot provide an exhaustive discussion of all these aspects here. However, it is important to note that philosophical approaches can only address  - in a consequent way -  explicit facts and knowledge that is represented with language-based argument, whereas for analyzing implicit representations in neuronal systems one needs different methods, too (mathematical or computational modeling), not offered by philosophy.

Partially levering from philosophy and linguistics is semiotics \cite{chandler2007semiotics}, with its emphasis on the relation between meaning and signs. Here the origins of meaning as well as  questions on syntax, semantics, and the usage of both are addressed. Thus, semiotics is instructive for the formal modeling of the acquisition of meaning. We have already explained above the triadic structure of Peircean semiosis, involving Object, Sign and Interpretant \cite{peirce1974collected,sowa2000ontology}, which is also part of the basis of our study. We follow these views, because the relation between the Object and its interpretation by an agent, the right side of the triangle, is of primary importance here.

Psychologists experimentally test hypotheses about the categorical and conceptual organization of the human mind. Rosch \cite{rosch1973internal} found out that for humans not all representatives  attribute equally to a particular category  (e.g., apples are considered to represent the fruit-category better, than plums or oranges). This is different from the classical view of philosophers stating that all representatives of the same Concept are equal. Psychologists, thus, adhere to the Prototype Theory that states that Concepts are representations whose structure is based on statistical analyses of properties of their \textit{members} \cite{laurence1999concepts}. Another frequently used conjecture is that Concepts are mental representations embedded in structures called theories. Thus, Concepts need to be analyzed in relation to other Concepts. Psychology analyzes how such theories are built during ontogeny \cite{carey1999knowledge}. By now evidence exists that abstract Concepts do have associations to sensory processing \cite{harpaintner2020grounding,troche2017defining}. Mirror neurons are potential candidates of conceptual representation at a fine scale \cite{rizzolatti2004mirror,craighero2007mirror}.

Biolinguistic theory investigates the course of evolution (and evolutionary leaps in particular) in attaining attributes allowing complex conceptual organization in the brain as well as capabilities to externalize that, using language. However the current line of thought here is that animals already have rich combinatorial thought, even though externalization might first be rather action- than language-based \cite{jackendoff2011human,fitch2017externalization}, which matches our approach taken in this study.

In robotics, the usage of the term ``Concept'' is rather new and not yet standardized \cite{kalkan2014verb,olier2017re,taniguchi2018symbol}. Taniguchi et al. \cite{taniguchi2018symbol} introduce some more clarity for the usage of the terms Category, Concept, etc. and suggest that Concept is an internal representation ``that goes beyond the feature-based representation of Category'' and that the semantic \emph{essence} of the Category needs be extracted to arrive at a Concept.

\underline{Cognitive entities in artificial neural networks.} Some works addressing higher cognition using neural networks also exist, but these systems have a different aim and have almost no intersection for comparison with our study. Hummel and Holyoak  \cite{hummel2003symbolic} developed a network-based cognitive architecture for acquiring schemas up to an ability to process relational representations, this way addressing human-like cognition \cite{penn2008darwin}. However, their architecture does not reach down to the sensory level and thus, does not perform grounding of the representations. Sandamirskaya et al. \cite{sandamirskaya2013using} develop neural representations in sensorimotor space based on dynamic field theory, which show emergent behavior that has qualities of higher order cognition. However, these systems are developed for demonstrating simple behaviors in a grounded way, which are meticulously pre-trained and, in addition, they need representations spanning the entire input and output range (e.g., all possible colors and space coordinates as well as all gaze directions), which makes representations relative bulky. Tani \cite{tani2014self} associates behavior and language learning in a continuous neuro-dynamical system trained in a supervised manner using pre-defined error back-propagation schemes. Similar to that, Olier et al \cite{olier2017re} use variational recurrent networks for associating different modalities to obtain situated behaviors of an agent. Xing et al. \cite{xing2018perception} developed a network mimicking different brain areas where integration of different sensory modalities for Concept acquisition is the main target. Possibly the study closest to ours is by Ozturkcu et al. \cite{ozturkcu2020high} investigating if neurons with meaningful (for a human) representations could emerge in small neuronal networks in the course of sensory-motor reinforcement learning. However, the semantics of the developed representations for the agent itself had not been addressed. Taken together, neuronal implementations grounded in sensory inputs and clearly targeted at helping to pin-down the concept of Concepts do not seem to exist in the current literature.

\begin{figure*}[ht]
\centering \includegraphics[width=17cm]{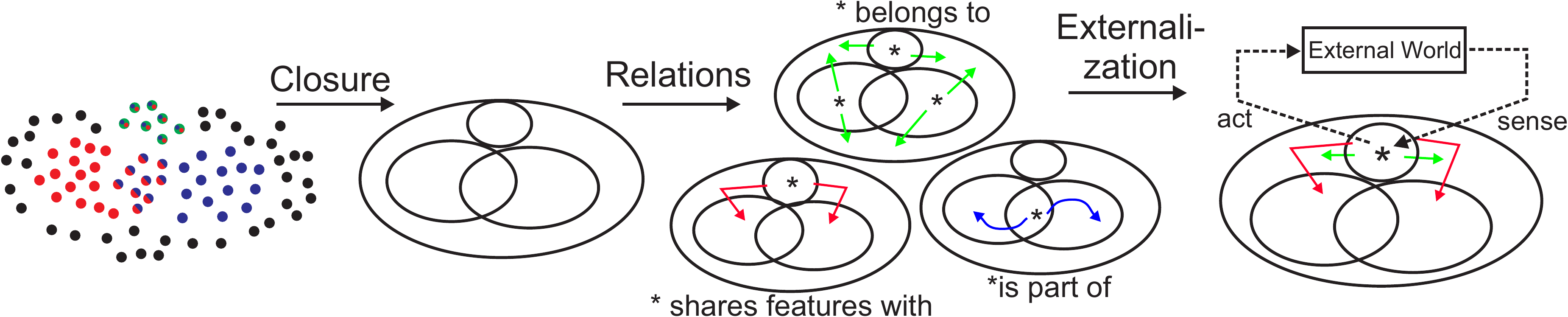}
\caption{Processes that define the transgressions from Categories to Concepts.}
\label{cat_con}
\vspace*{-4mm}
\end{figure*}

\subsection{Definitions: Categories and Concepts}

From the above  it is evident that there is still no neuronally-implementable way existing to arrive at Concepts. To approach this, we will first discuss how we would see the difference between Category and Concept and present a view that allows for partial implementation.

Following \cite{taniguchi2018symbol}, we posit that Categories are un-reflected clusterings within a sub-symbolic feature space, whereas Concepts are the narrative, the rule, the essence of the thereby captured entities. As humans we can speak out about (i.e. externalize) concepts using language. E.g. we say:  this Concept is A and not B, it contains C but not D and it relates to E, etc. These narratives or ``rule-sets'' for a Concept are in essence relational and, thus, clearly go beyond mere ``feature constellations'', which you could obtain by clustering. The aspect of ``relations'' is, thus, in our opinion essential for Concept and we will build our model of Concept formation based on the following definitions (see Fig.~\ref{cat_con}, examples are given below to make this clearer):
\begin{itemize}
    \item Categories can be understood as clusters. They are defined only through (often pairwise) local similarity operations between their members and are often represented by a central exemplar.
    \item
    Concepts are formed by relations and Categories are used as the basis to form them. In detail:
    \begin{itemize}
    \item
    \emph{nascent Concept: }The first relation that is needed to bootstrap Concept formation is closure (i.e., the inclusion vs. exclusion operator) obtained by drawing the boundary around clusters.
    \item
    \emph{consolidated Concept:} Concepts are consolidated by relating them to other Concepts.
    \item
    \emph{externalizeable (final) Concept:} Finally Concepts must entail some kind of agent-owned explicit-ness. The consequences of acting according to a Concept need to be observable (or self-observable).  Arguably, self-observation is one central step for the forming of a mental image that represents then ``this'' Concept.
    \end{itemize}
\end{itemize}

\subsection{Examples}

\underline{Categories as clusters}: Let us consider a set of vectors in RGB space. One could now use an algorithm that sorts this set according to statistical correlations between the vectors into a map-like representation. For example, this would work with the Self-Organizing Map (SOM) algorithm \cite{kohonen1990self}. Similar colors would this way be placed near to each other with continuous transitions from one to the other. The algorithm would this way define clusters that can be seen as color-categories. But it would not be able to tell that ``this is red and this not''.


\noindent
\underline{Nascent Concepts via closure:} Closure is fundamentally needed to be able to formulate a relation that says this is red and this is not red. Hence, you need to (mentally) draw a boundary enclosing all points according to your idea of which color is (still) red and which not.
This is the first step towards Concept and requires the non-local process of ``drawing this line'', i.e., defining a decision boundary.

\noindent
\underline{Consolidated Concepts from using more-complex relations:} Closure can only create a narrative that says ``belongs to'' versus ``does not belong to''. Human concepts are usually far deeper than that and rely on many more relations. For example, more complex relational operations would in this case be: [a (traffic light) is (composed of)  (green), (yellow) and red] or  [red is (on top of) the (traffic light)]. Where [~] refers to the relation used to explain the Concept ``red'' and (~) refers to other Concepts underlying these respective relations. Clearly, entities in (~) are Concepts in their own right and would have to have their own explanations without which [~] would be incomprehensible. Here, we must limit ourselves to addressing simple relations at the level of propositional logic as our system cannot attempt explaining more complex human concepts.


\noindent
\underline{Externalizeable (final) Concept by agent-owned explicitness:} Representations within a brain (or artificial neural network) that capture all aspects above but just exist without triggering any behavior are -- at best -- ``incomplete'' Concepts. Concepts need to entail behavior in a closed loop context. Note that human mental simulation is here subsumed also under ``behavior''. Also note that externalization comes for humans in several stages. E.g., children at the age of about three can correctly respond to the request: ``Please give me the red pencil!'' but they cannot yet answer the question ``Which color does that pencil have?'' (in spite of using, in other situations, the word ``red''!). They have, we would say, only a partially externalizable Concept of ``red''. While they must be able to form a mental picture of ``red pencil'' to be able to react to the request, the final narrative stage (discourse or self-discourse) has not yet been reached at that age.



This allows us to consider how an artificial neural system could arrive at Concepts. We would need neurons that perform (1)  \emph{closure} and represent (2) \emph{relations} between entities in their synaptic weights. Ideally the system should then (3) \emph{externalize} the Concept and create behavior and finally a narrative. The system presented below will show properties 1 and 2 albeit only for some simple cases. Property 3 will emerge, too, as the system will produce responses (after Concept formation) for ``imagined situations'' as if it were forming some kind of mental image, like our 3-y old child from above. But narratives are not yet being made by this system.

Clearly, our system will not even get close to the complexity of a 3-y old. However, we believe that the aspects shown below, all of which are biologically realistic and generalizable (independent of input modalities), may help to plant the cognitive processes for Concept formation better in some algorithmically sound and transferable soil.

\begin{figure*}[ht]
\centering \includegraphics[width=18.5cm]{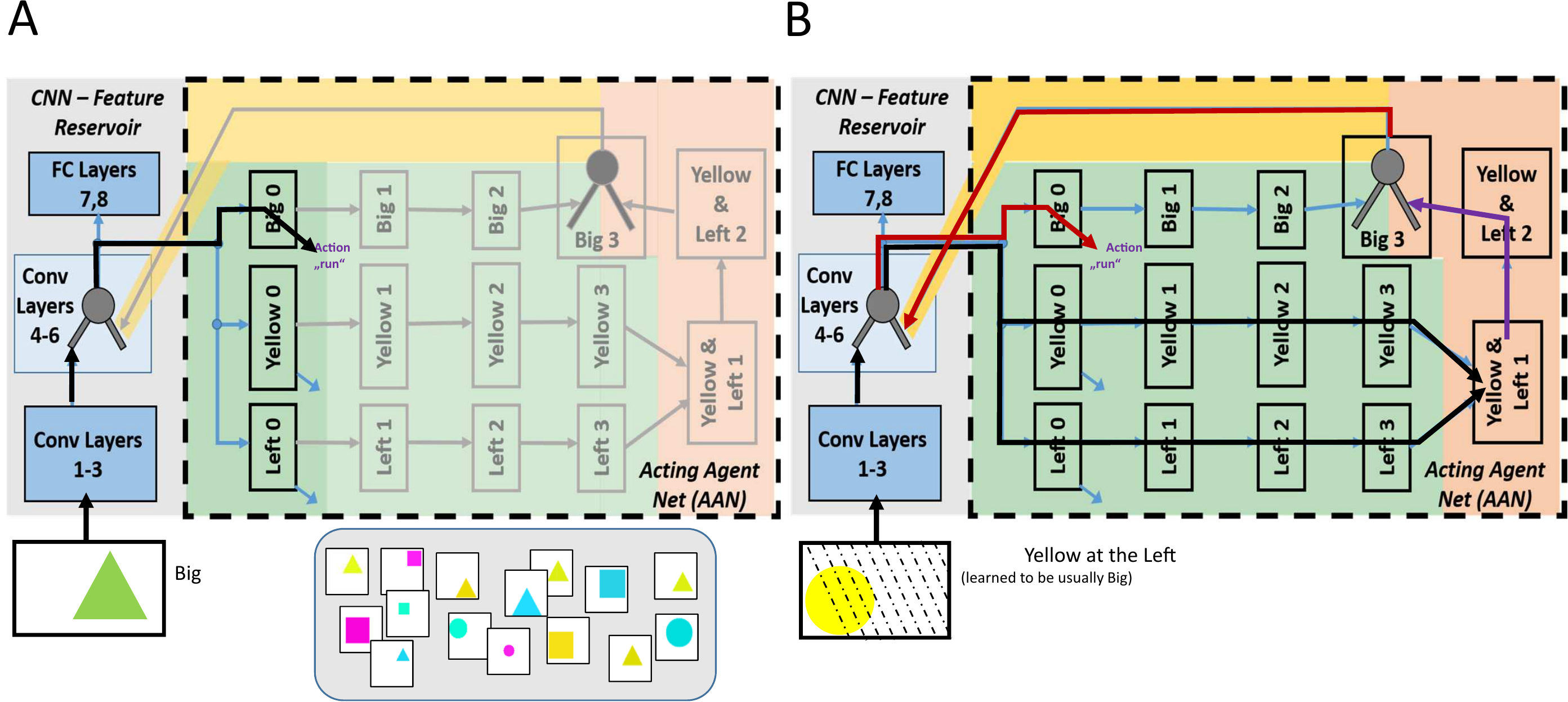}
\caption{Architecture of the system and signal flow. Convolutional layers 4-6 of a pre-trained neural network are used as feature reservoir (grey box) for training network on the right (dashed box) using associative learning. The small neuron icons in Big 3 and in the convolutional layers indicate that these neurons use 2 dendrites each for learning. A) Feed-forward: After training, the network in the dark green box triggers the action ``run'' of a virtual agent in case the object in the image is big. B) Feedback: A small yellow item on the left (which is really an occluded big one) triggers ``running'' via feedback. Fat arrows: black=feed forward, violet=feed forward after AND operation, red=feedback. Note, the action in our system is virtual, represented merely as a signal for ``run'', which is used for evaluation of the system. Inset in A shows examples of input images.}
\label{scheme}
\end{figure*}

\section{Methods}

\subsection{Basic Model Structure and Goal - Overview}
The goal of our network is to emulate an agent that should learn to react in an appropriate way to certain Concepts, which arise from visual signals.

To achieve this, the systems consists of two fundamental components. There is a conventional convolutional neural network (Fig.~\ref{scheme} A, B left, grey) pre-trained once in a supervised manner on a surrogate task and used as \emph{feature reservoir}. This could be considered related to the early visual areas (e.g., V1) in the visual cortex \cite{eickenberg2017seeing}.

In addition to this, we implemented a second network, called Acting Agent Net (AAN, Fig.~\ref{scheme} A, B  in the dashed box). This network is indiscriminately fed features from convolutional layers 4-6 of the CNN and is trained using associative learning. First, we use  reflex avoidance learning, which allows selection of environmentally relevant input signals (and, thus, delivers reinforcement from the environment). In the following stages, we use variants of classical Hebbian learning to process associations between those environmentally-relevant signals.

The primary operation of the network is to trigger a reflex (follow black arrows in Fig.~\ref{scheme}~A) by which the agent responds with the virtual action ``run'' in a feed forward manner, if an image with a big shape (in human terms: a dangerous entity) is processed by its early sensorial layers. The central task of the network, however, is to replace this primary reflex by learning a --- in the inputs existing --- (hidden) conjecture, namely that yellow entities on the left are usually big and, thus, to learn [Yellow \& Left $\rightarrow$ Big] and react accordingly.

The fat arrows in Fig.~\ref{scheme}~B show how to augment the primary operation in the AAN by feedback to react to  this conjecture. Briefly (for details see ``Specific Architecture and Mode of Operation'', below):  First an intrinsic \emph{closure} process on the different visual features leads to nascent Concepts (green box). The agent then learns that certain \emph{relations} between these nascent Concepts exist (light brown box), allowing it to form consolidated Concepts, which are indicative of a big entity. If that happens, feedback (dark yellow box) to the sensorial levels is provided ``as if'' a big entity had been visible. The running action is then triggered by this ``imagined'' big entity, indicative of a final, \emph{externalized} (albeit pre-linguistic) Concept.


\begin{table*}[ht]
\begin{center}
\caption{Data sets.}
\label{table_datasets}
\begin{tabular}{|l|l|l|}
\hline
Data set                                                        & Size  & Remarks                                     \\
\hline
\begin{tabular}[c]{@{}l@{}}CNN pre-training set~\\\end{tabular} & 24300 & 27 x 900 samples: 27 = 3 sizes x 3 shapes x 3 colors, no Medium or Small Yellows on the Left.                                            \\
\hline
AAN training set                                                & 2742  & \begin{tabular}[c]{@{}l@{}}27 x 100 samples + 6 x 7 samples*: 27 = 3 sizes x 3 shapes x 3 colors, no Medium or Small \\ Yellows on the Left; 6 = Medium x 3 shapes + Small x 3 shapes, all Yellow on the Left. \\Samples generated independently from the CNN pre-training set. ~ \end{tabular}      \\
\hline
Baseline test set                                               & 2700  & \begin{tabular}[c]{@{}l@{}}27 x 100 samples: 27 = 3 sizes x 3 shapes x 3 colors, no Medium or Small Yellows on the Left.\\Samples generated independently from the training sets.   \end{tabular}                                                      \\
\hline
nBYL test set                                                   & 600   & 6 x 100 samples: 6 = Medium x 3 shapes + Small x 3 shapes, all Yellow on the Left.                                                                 \\
\hline
\multicolumn{3}{l}{\begin{tabular}[c]{@{}l@{}}*\textit{To analyze the influence of the training set composition, as shown in Fig.~\ref{rule_viol} we also used AAN training sets with variable}\\\textit{ number of Small and Medium Yellows on the Left (none, 6x5, 6x7, 6x10, 6x20, 6x30, 6x40, 6x50, 6x75 and 6x100).}\end{tabular}}
\end{tabular}
\end{center}
\end{table*}

\subsection{Data set}\label{section_dataset}
For this study we use simple computer-generated RGB-images with $100 \times 100$ pixels, depicting uniformly colored shapes with three pre-set features: geometric shape (circle, square, triangle), color (yellow, magenta, cyan) and size (big, medium, small). This creates 27 possible combinations. Some examples are shown in the inset in Fig.~\ref{scheme}~A. colors and sizes are variable within pre-defined non-intersecting intervals. These shapes can be placed anywhere on the canvas.

First we train the CNN using a set of $24300$ images in total ($900 \times 27$). This is done in a standard supervised manner only once and the CNN remains unchanged for all experiments. The output of the CNN is then able to recognize all 27 picture-types, but we do not use the information in the last fully connected layers of the network. Instead, we use the feature information present in convolutional layers 4-6 of this network as acquired during CNN training.

For the AAN, the training set consists of $2742$ such RGB images. Apart from the explicitly introduced features (color, shape, and size) one additional \emph{implicit} feature: Left is used in this study. Feature Left defines the fact that pixels of the given shape enter the left $30$ pixel wide rectangle on the canvas, where for non-Left images this area is empty. We use $100$ images for each shape $\times$ color $\times$ size combination ($=2700$). One specific additional aspect of this data set is that yellow images on the left are most of the time big. Thus, all yellow shapes in the aforementioned 2700 are generated so that small and medium size objects are never ``on the left'', while the other colors and sizes are distributed over the canvas uniformly. Thus, this is a hidden statistical regularity in the agent's \emph{world}, from which it should draw appropriate conclusions. We use $42$ non-Big Yellow shapes of all kinds ($6$ kinds $\times$ $7$) ``on the left'' to add randomness (15\%) to the statistical rule [Yellow~\&~Left $\rightarrow$ Big], which reflects better any real world situation where rules are most of the time not-perfect.

For testing of the AAN-system we use two data sets:

1) $2700$ images (shape $\times$ color $\times$ size $\times$ $100$) generated using the same procedure as for the training set, \textit{without} non-Big Yellows ``on the left''. We call it \textbf{baseline} test set.

2) $600$ images of all possible non-Big, Yellow shapes ``on the left'': non-Big that is, small and medium circles, squares, and triangles, $100$ each (called \textbf{nBYL} test set). Information on all data sets used is summarized in Table~\ref{table_datasets}.

\subsection{Specific Architecture and Mode of Operation}

The system is activated by presenting images one after the other at the input. Hence, we operate in discrete time chunks. The activation from the image spreads through CNN layers and from there excites all blocks of neurons in the AAN one after the other as indicated by the thin arrows in Fig.~\ref{scheme}, and may elicit some kind of (virtual) behavior. This mode of operation is found in a similar way in many mammals, where an input may trigger a certain motor program, which continues for a while, until accumulation of information from the next input(s) triggers a change in behavior \cite{summers2009current}.

As mentioned above, the AAN network (dashed box) uses and evaluates intrinsic information contained in the convolutional layers 4-6 of the CNN.

Layer 0 defines the basic behavior of the agent. Importantly, as we do not simulate any real agent all behaviors are only virtual. For example, if pool Big 0 in layer 0 (Fig.~\ref{scheme}~A, top) is activated above threshold (see details in the Appendix), a virtual behaviour called ``run'' will be elicited. For all zero layer pools, we use the Input Correlation Learning rule (ICO) \cite{porr2006strongly}, which is a form of associative learning based on stimulus substitution in classical conditioning \cite{Barto1982simulation}. For that, the agent has to have a pre-wired reflex.  If some sensory input in a persistent way arrives before the reflex-triggering signal, the agent learns to act on detecting that earlier stimulus, thus, learns to avoid the reflex. Thus, the reflex-triggering signal acts as an error term that drives the learning. This is associated to the function of the Cerebellum (reflex avoidance learning, see review by Dean and Porrill, \cite{dean2014decorrelation}). Hence, for all zero layer pools, we have pairs of ``reflex-triggering'' and ``earlier'' input signals. For ``run'' the reflex-triggering signal is a  dangerously approaching shape, hence: before learning ``looming'' triggers run as reflex. However, before it approaches, an earlier signal exists, which is that it is big, hence: after learning ``big'' triggers run proactively. As mentioned above, all ``movements'' in our model are only virtual, hence the big object's approach (``looming''), too. Yellow and Left are set up in the same way but we do not analyze their stimulus input pairs explicitly.



Activity established in layer 0 travels to layers 1-3 and further. Different from layer 0, all other layers use variants of classical Hebbian learning, thus do not use error terms (see Appendix for details). While Layer 0 extracts categories, layers 1-3 iteratively sharpen (discretize) the representations attained in layer 0 for Big, Yellow, and Left.


Above we had stated that the hidden conjecture of this world is that most items, which are yellow and ``live'' on the left side, usually are also big. Hence, it would be useful for the agent to learn the conjecture [Yellow~\&~Left $\rightarrow$ Big] and react with running. Consider a scenario in the real world: it will be very useful for you to realize that this yellow (color feature), partially occluded agent at the water hole (location feature) very likely is a tiger from which you should run away, not waiting to see the whole big (size feature) beast. Thus, two more layers exist for this AND operation (light brown box, violet arrows). Yellow~\&~Left 1 learns the AND, and Yellow~\&~Left 2 is another layer for sharpening the results that came from Yellow~\&~Left 1.

The association [Yellow~\&~Left $\rightarrow$ Big] is formed in the neural population Big 3. This group of units receives multi-modal input from Big 2 as well as from Yellow~\&~Left 2. To allow learning of such multi-modal inputs, we use a simulated dendritic structure with two branches each, where synaptic plasticity at each branch operates independently of the other branch. This type of local branch specific plasticity is known from many neurons \cite{cichon2015branch} and represents a powerful mechanism for non-destructive signal structuring in the brain. In our study, dendritic structures provide the maximum of the two branches to the output (thus, are active in case at least one branch is active).

Finally the signal from Big 3 travels back to the convolutional layers 4-6 in the CNN (dark yellow box). Hence, these layers also receive two different types of inputs, on the one hand from the lower convolutional layers and, on the other hand, from Big 3. The same dendritic structure as described above is used here, too. These feedback connections (red arrows, Fig.~\ref{scheme}~B) can now activate the behaviour ``run'', in case this behaviour has not yet been activated in a feed forward manner. Hence, this will happen (after learning) as soon as the agent sees a yellow item on the left.


\begin{figure*}[ht]
\centering \includegraphics[width=17cm]{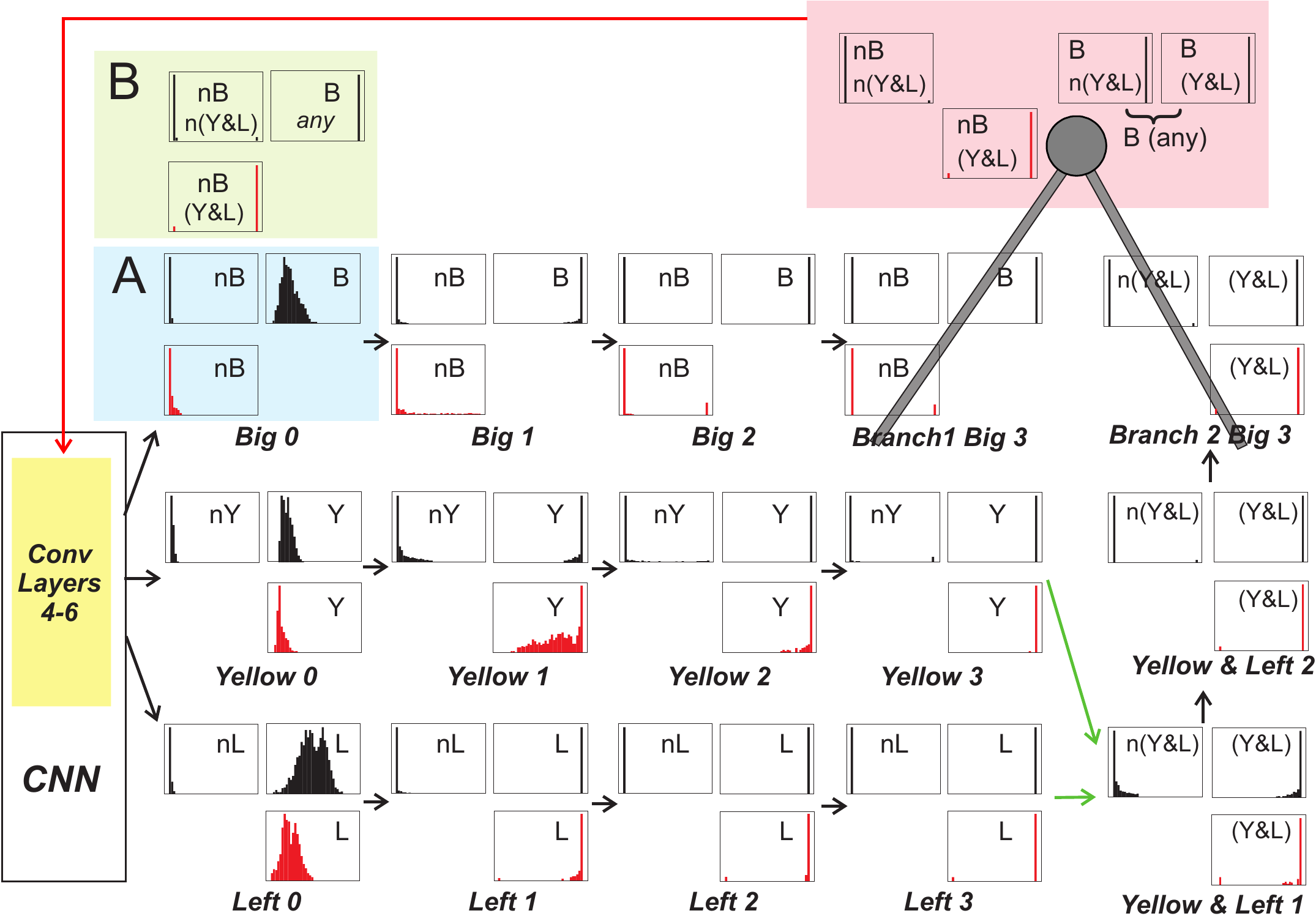}
\caption{Histograms of average pool activations for \textbf{baseline-} and  \textbf{nBYL} test set (black vs. red). The \textbf{baseline} test set is similar to the training set, while the \textbf{nBYL} test set is composed of non-Big Yellows on the Left. Histograms are given in the same order as in Fig.~\ref{scheme}. Inputs into the two dendritic branches of  Big 3 are shown overlaying the dendrite icons. The result of the computation performed by Big 3 is shown in the pink box. Activation of Big 0 before feedback is shown in the blue box (panel A, top left) and after feedback in the yellow box (panel B). Abbreviations are B: Big, nB: non-Big, Y: Yellow, nY: non-Yellow, L: Left, nL: non-Left, Y$\&$L: Yellow \& Left, n(Y$\&$L) non-Yellow \& Left. Simulation parameters are provided in Tables \ref{combi-tab} to \ref{table_params_balanced_hebb} in the Appendix.}
\label{histos}
\end{figure*}

We use $N=300$ neurons for each block, except for the CNN-activation block, where $4416$ neurons are present  and we usually use $50$ connections from the previous block to the next block, which are chosen randomly in a uniform manner with very small starting weights. Only the connections to Yellow~\&~Left 1 are $25$ each, leading to 50 connections in total again. Feedback is organized so that each neuron from pool Big 3 provides connections to the CNN-activation neurons in the convolutional layers 4-6.

All neurons have sigmoidal activation functions except in the CNN layers where ReLU activations are employed. Detailed equations for the network activation and learning rules are provided in the Appendix.

\section{Results}

To illustrate how the system works, first we show in Fig.~\ref{histos} one example of the activation patterns in the network obtained after learning. Activation patterns for each pool of 300 neurons are provided as averages, plotted as histograms over all images for the \textbf{baseline} test set (black, this is a training-set-like set with 2700 images) and the \textbf{nBYL} test set (red, contains 600 non-Big Yellows on the Left). All x-axes are scaled between 0 (no neuronal activation) and 1 (maximal activation). The desired outcome, as seen in the histograms, is that the presence of a feature, for example Big (B), in the corresponding pools (Big 1, 2, or 3), should lead to high activations, the absence (nB) to low activations.

In the initial layers 0,  histograms are relatively wide with narrow separation between two categories (e.g., Big, B) vs. non-Big, nB). In layers 1, 2, 3 increasingly stronger discretization is present. The Hebbian learning performed in these layers leads to a continuous sharpening of the distributions without any additional mechanism. This is instrumental for being able to perform the AND operation successfully (see Yellow~\&~Left~1). The result of the first AND operation is then one more time sharpened (Yellow~\&~Left~2) and transferred to Branch 2 of the dendrite. Here learning commences and creates locally the distributions as displayed on that branch.

The other branch receives the output from Big 2 and also performs learning. Results of both dendritic learning processes are then combined at the somata of the Big 3 neurons by a maximum operator. Note that a maximum operator corresponds for well discretized signals to performing a logical OR operation common for many neural systems when operating at a low firing threshold.  Thus, the output of Big 3 is close to one in case the shape is big, OR if the shape is Yellow~\&~Left as shown in the histograms in the pink box above. This signal is fed back into the feature reservoir (convolutional layers), and modifies those through a dendritic structure similar to the one described above. Here one branch  receives the original activation from the CNN, while the second branch receives the feedback.

It is known from all deep learning networks that the activation of individual nodes is hard to interpret. This is here true, too, and feed forward as well as feedback activation in the CNN layers is fairly non-descriptive and quite dispersed (not shown). Of essence, however, is that these activations will converge ``in the right'' way to layers 0 and here specifically to Big 0, which triggers the running action.

Hence, the CNN provides two types of inputs to layer Big 0. If a big item appears in an image the feed forward pathway (Fig~\ref{scheme}, A) will trigger the action ``run'' and we can assume that this leads to the immediate disappearance of the image from the sensorium (as the agent has turned around). Re-excitation of Big 0 via the feedback loop will have no consequence because the running action will have already commenced in this case. Alternatively, if a non-big, yellow item appears on the left side of the image, Big 0 receives  the feedback signal (Fig.~\ref{scheme}, B) which makes the agent run. These two cases are shown in Fig.~\ref{histos} panels A (blue - feed forward activation) and B (yellow - feedback activation).

Most importantly, in the feedback case the excitation for the set \textbf{nBYL} has changed from close to zero (panel A, red histogram), to close to one (panel B, red histogram). Thus, feedback induces the effect, that even non-big yellows on the left strongly excite the pool Big 0, which is responsible for the action ``run'' in the model system.  Statistics quantifying the discretization effects as well as about the logical conjectures, seen in Fig. \ref{histos} are presented in the Supplementary Material and prove that the case we are showing was not cherry-picked. In addition, we show in the Supplementary Material for a control case where a Magenta input is used instead of Yellow, that no strong signal is produced in Branch 2 Big 3. This is because the learning rule (Balanced Hebb, see Appendix for details) used for Branch 2 Big 3 does not increase weights in case a non-existing conjecture is probed, which here is [Magenta \& Left $\rightarrow$ Big].

\begin{figure}[ht]
\centering \includegraphics[width=6.5cm]{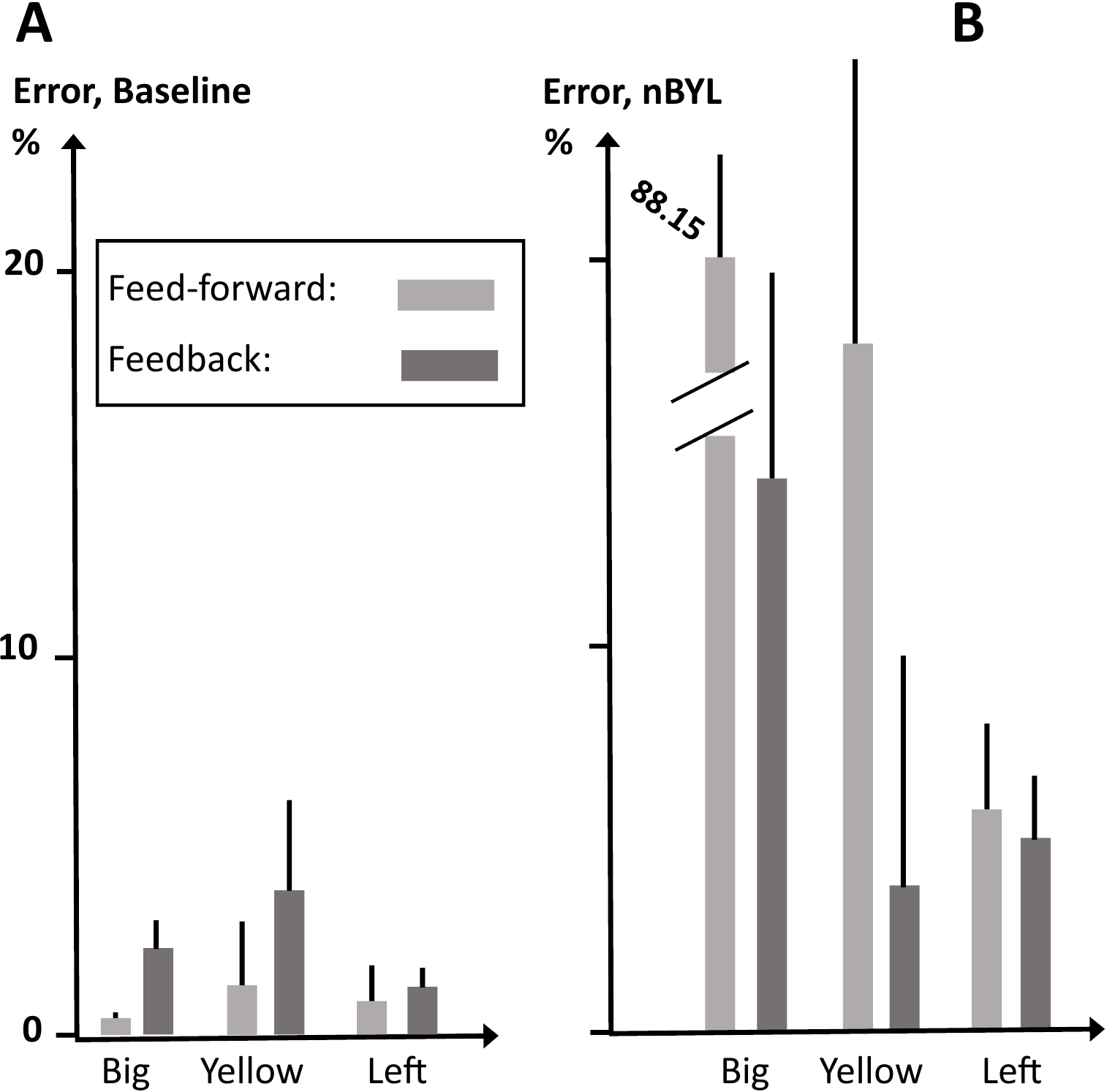}
\caption{Error rates for 10 runs with and without feedback for both test sets, (A) \textbf{baseline} and (B) \textbf{nBYL}. Error bars show standard deviation.}
\label{error}
\end{figure}

Next we analyse the errors of the system obtained in feed-forward mode (Fig.~\ref{scheme}, A) as compared to the ones obtained after feedback (Fig.~\ref{scheme}, B). We show this for the two data sets \textbf{baseline} and the  \textbf{nBYL} for layer zero. Our aim is to induce, using the feedback, the action ``run'' for the test set \textbf{nBYL} without increasing the error rate for the action ``run'' on the test set  \textbf{baseline}, where the error rate is close to zero without feedback.

Averages of error rates and standard deviations for 10 trials for the two cases (feed-forward and feedback, see Fig.~\ref{scheme}) are provided in Fig.~\ref{error}. For the most relevant test set \textbf{nBYL}, it can be seen that the error for Big (=``run'') drops substantially when feedback exists (Fig.~\ref{error}~B, from about 88\% to 14\%). There is a very small increase in errors for the test set \textbf{baseline} (Fig.~\ref{error}~A). This is due to the fact that in forward mode up to layer 3 still small errors do exist in the discrimination between feature vs. non-feature. For example, look at the histogram nY in layer Yellow 3 in Fig.~\ref{histos}, where you can see a small peak of neurons with high activation. These small errors are fed back and lead to minor deterioration of the signal in the convolutional layers and finally to the here observed small increase in error in layers 0 after feedback (Fig.~\ref{error}~A).

Summarizing, after feedback, we have $2.23\%$ error for action ``run'' for the \textbf{baseline} data set  and $14.27\%$ error for the  \textbf{nBYL} data set. The increase in the error for correct behaviour in the \textbf{baseline} data set is only a couple of percents, while the decrease in the error on \textbf{nBYL} data set is massive, reduced by $73.88\%$. Thus, all in all feedback brings new advantageous properties to the system without much disturbing performance of the initial feed-forward system.

It is, however, non-trivial to characterize the feedback as such. Feedback in our system modifies the activation of the convolutional layers, neuron by neuron, but we had argued above, that it is hard if not impossible to interpret individual activations in the convolutional layers.

However, we can ask, how ``similar'' is the feedback activation to the original activation and one way to show this is by calculating the correlation between feedback and feed-forward activations in the CNN. We use the following procedure using the \textbf{baseline} test set: align the feed forward activations of the neurons in convolutional layers 4-6 into a one-dimensional vector
$\bm{x^c(k)}=(x^c_1(k),x^c_2(k),\dots ,x^c_{4416}(k))^T$ for images  $k=1,2,...,2700$ and in the same order align the feedback coming to the corresponding neurons in the convolutional layers  $\bm{x^{fb}(k)}=(x^{fb}_1(k),x^{fb}_2(k),\dots ,x^{fb}_{4416}(k))^T$, thus obtaining the second vector. Then we calculate Pearson's correlation coefficient. This can either be done directly between the two vectors $\bm{x^c(k)}$ and $\bm{x^{fb}(k)}$ or after averaging across a set of same-feature images (e.g., all big images): $\bm{\bar{x}^c}$ and $\bm{\bar{x}^{fb}}$. Results are provided in Table~\ref{tab_cor}.

At first we note that image-by image correlations are far away from one, indicating that different images of the same category would (as expected) activate the convolutional layers in quite different ways. This also shows that there is not \emph{one} specific image or any small set of images the feedback would ``hook onto'' and predominantly represent. Instead, the average of activation in image sets tell the story. Here we obtain a very high correlation coefficient of $0.96$ for Big and, thus, can state that the average feedback signal is similar to the average feed forward activation of convolutional layers obtained by big shapes. For comparison, we show in Table ~\ref{tab_cor}  correlations for Medium and Small shapes, too, where those correlations are much smaller.

\begin{table}[ht]
\begin{center}
\caption{Correlation coefficients between original activation and learned feedback for different shape sizes in \textbf{baseline} data set.}
\label{tab_cor}
\begin{tabular}{llllllll}
Correlation type  & Big & Medium &  Small \\
\hline
Image by image (mean $\pm$ STD)

 & 0.45
 & 0.16
 & 0.06
\\
~

 & $\pm$ 0.11
 & $\pm$ 0.16
 & $\pm$ 0.16

\\
Average of activation in image sets & 0.96 & 0.36 & 0.08\\
\hline
\end{tabular}
\end{center}
\end{table}

\begin{figure*}[ht]
\centering \includegraphics[width=17.5cm]{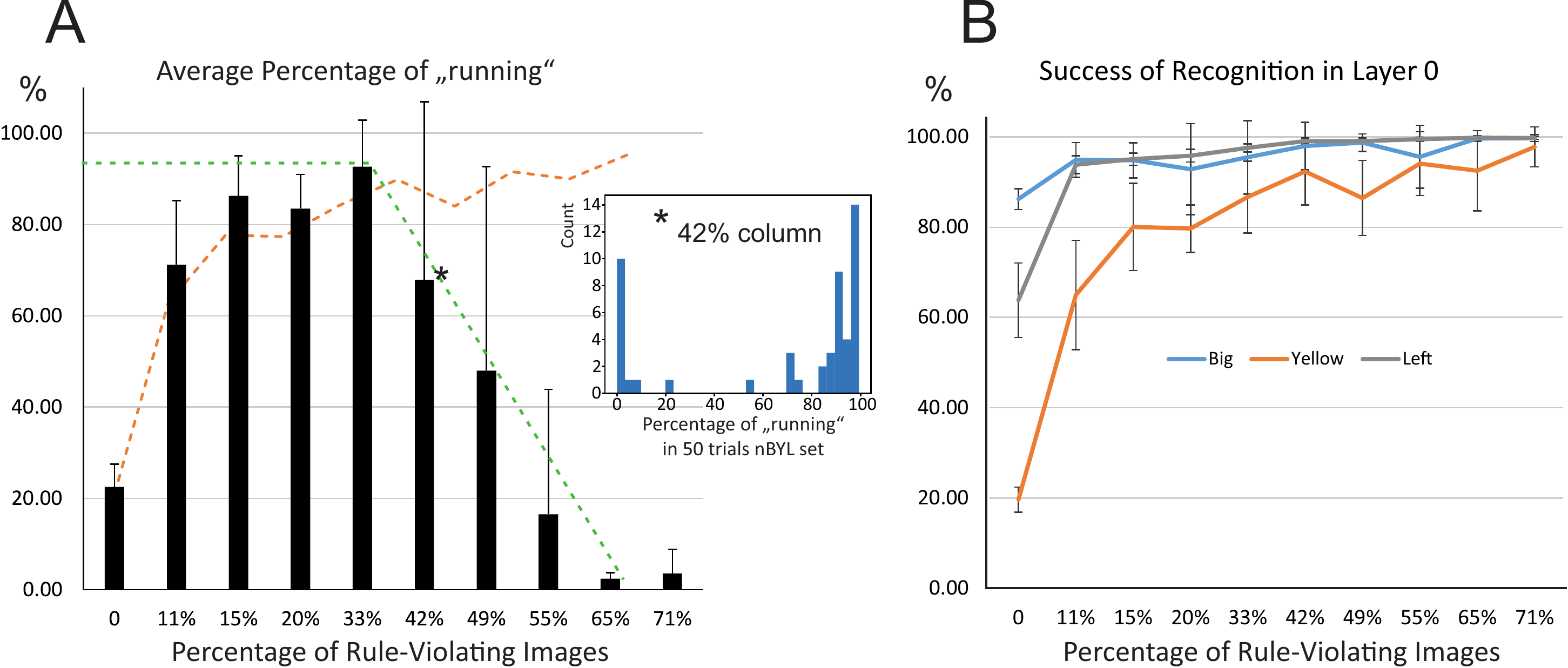}
\caption{Behavior of the system when increasing the fraction of rule-violating images in the training set, evaluated on nBYL test set. Data based on 10 trials each. A) Average percentage of running. Inset on experiment with 50 trials and 42\% rule-violating images renders a bi-modal distribution. B) Success of feature recognition in layers 0 for the three basic features.}
\label{rule_viol}
\vspace*{-2mm}
\end{figure*}

Different from this, it is, however, important to better understand to what degree the system can extract the ``hidden rule'' that yellow items at left are (almost) always big. So far in the training set we always had 15\% of yellow images on the left that were violating this rule: a total of $288$ yellow images exist on the left, of which $246$ are big and $42$ not. This produced an error of about 14\% in ``run'' in nBYL test set (Fig.\ref{error}~B). In Fig.~\ref{rule_viol}~A we show the average percentage of running in nBYL test set when adding more and more images  that violate the rule to the training set, call them outliers. Up to a fraction of 33\% outliers running behavior improves, above 33\% system produces less and less running. For explanation of the imperfect behavior of the system at low fraction of outliers, we show in panel B, how well layer 0 can recognize the different features, where Yellow performs less good than the others, especially at low percentage of outliers. Hence, the orange curve is the limiting factor for the performance in (A) on the left side (see dashed copy of this curve). The drop on the right side in (A) is explained by the fact that towards the right the rule ``Yellow on the Left is most of the time Big'' becomes weaker and stops existing at a fraction of 50\% of ``outliers'', where the outliers cannot be called so anymore. Hence, this drop in running behavior is actually desired for the right side. The green dashed curve represents a rough estimate up to which fraction the system ``believes in the rule''. The inset histogram shows what happens in the transition range, where also the standard deviations in panel A are big. For 42\% outliers we receive a bi-modal distribution where the system sometimes still follows the rule with a high running percentage, but sometimes not.

\begin{figure}[ht]
\centering \includegraphics[width=9cm]{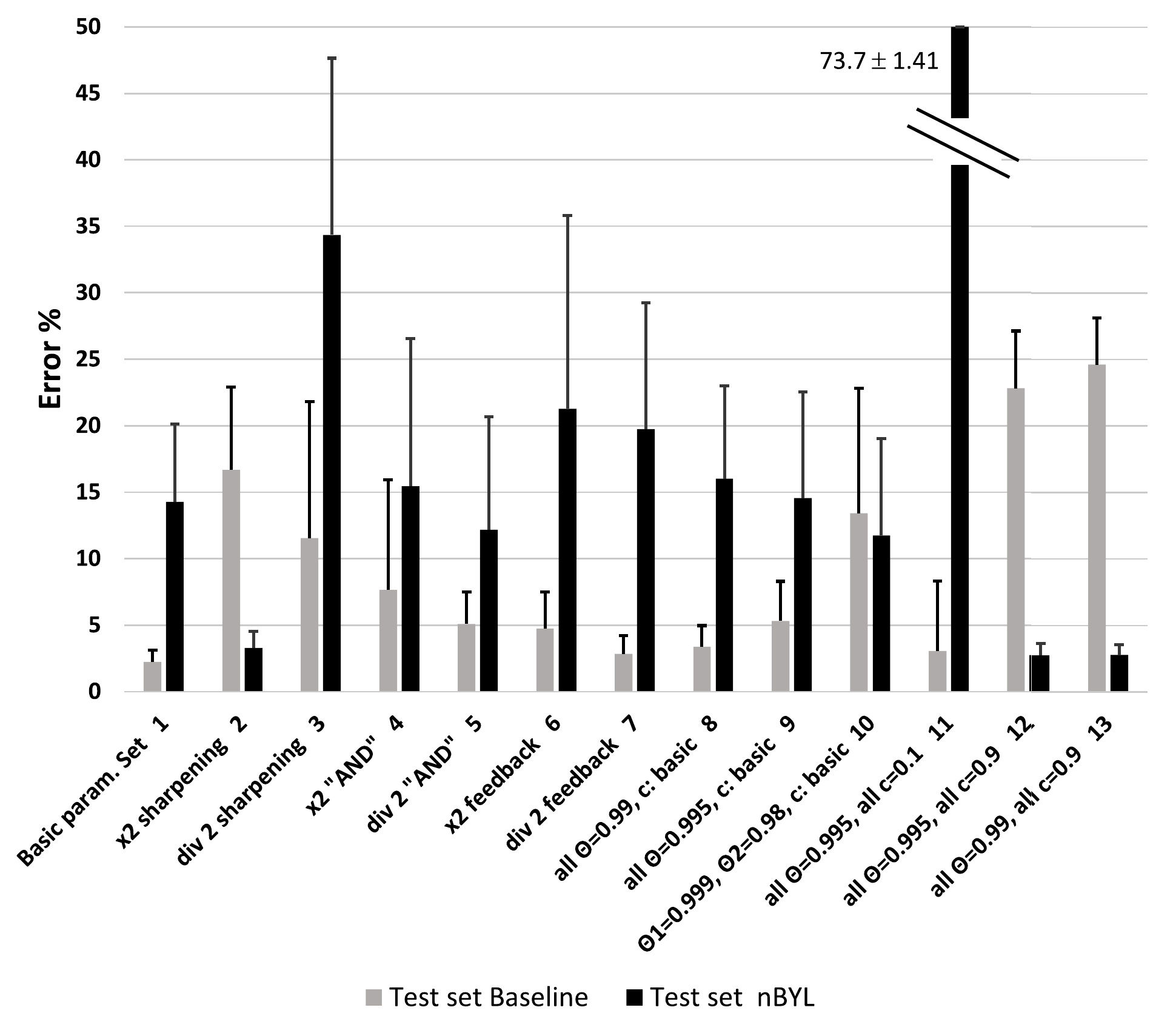}
\caption{Parameter analysis of the system. For explanations see text.}
\label{chart01}
\vspace*{-4mm}
\end{figure}

Fig.~\ref{chart01} shows results of a parameter analysis to demonstrate the robustness of our system. The system, we are presenting, depends on the learning rates in the different pools and on the parameters involved in annealing of the learning (see Appendix for detailed equations).

In Fig.~\ref{chart01} we show errors for \textbf{baseline} (grey column) and  \textbf{nYBL} (black column) test sets. Overall, the aim of any parameter combination would be to keep performance as good as possible on the \textbf{baseline} test set, while getting also the best performance on the \textbf{nYBL} set.

Column pair 1 shows the performance obtained with the parameter set which has been used for all experiments in this paper. The following comparisons show that this is indeed a good choice.

In column pairs 2-7 in this figure we show results when we vary the learning rates for different neuron pools: layers 1-3, responsible for sharpening (column pairs 2 and 3), ``AND'' pool (column pairs 4 and 5) and the feedback (column pairs 6 and 7). We increase (marked as x2 in the figure) and decrease (marked as div 2) the learning rates twice and evaluate the system on the two test sets after learning. The results show that errors on the two data sets vary to some degree when changing the learning rate, but the general operation of the system is not destroyed. The only larger increase in error happens when decreasing the learning rate for the sharpening subsystem (column pair 3). With increasing the learning rate for sharpening (column pair 2), the error rate for the test set \textbf{nYBL} drops, but then the error rate for the \textbf{baseline} increases, which is due to an ``over-training'' of the feedback.

To the right of this in column pairs 8-13 we show results when we use our standard learning rate but vary the parameters that regulate the annealing (``stopping'') of the learning. Annealing starts as soon as the output of a neuron is above threshold $\Theta$. Furthermore, we also vary parameter $c$ which regulates the speed of annealing, where a smaller $c$ brings faster annealing. In the standard network (column pair 1) we use softer annealing $\Theta=0.99$ and $c=0.9$ for the first sharpening layer (layer 1) and for the feedback, while elsewhere in the system abrupt annealing is used  ($\Theta=0.995$, $c=0.1$).

The final result, however, is not very sensitive to the change in threshold $\Theta$ (column pairs 8-10). However, the learning process is destroyed in case the annealing is too abrupt with $c=0.1$ everywhere (column pair 11).
For column pairs 12 and 13 we used a large value of c for all pools, which makes annealing slow, but this lead to large errors in the \textbf{baseline} test set, indicative of over-training.

However, overall, the system has a relatively wide parameter range where the performance reaches the goal of small errors on the \textbf{nYBL} data set, without destroying the performance on the \textbf{baseline} test set.

Finally, Table \ref{table_ablation} provides results of an ablation study. We connect the ``AND'' pool as well the feedback to layers \textbf{0}, \textbf{1}, \textbf{2}, or \textbf{3}, where the latter is the full system and measure  errors in the two test sets. Results show that errors get smaller with adding layers 1, 2 and 3

\begin{table*}[ht]
\begin{center}
\caption{Ablation study. Means and standard deviations of error are provided on the basis of 10 trials.  Asterisks show the variants where the change from the previous (smaller) variant of the system was statistically significant ($p<0.05$ for the t-test).}
\label{table_ablation}
\begin{tabular}{lllll}
 & Only Layer 0 & Layers 0-1 & Layers 0-2 & Full system \\
 \hline
Error in test set Baseline after feedback
 & 14.00 $\pm$ 3.86
 & 11.65 $\pm$ 5.65
 & 4.73 $\pm$ 2.78*
 &  2.23 $\pm$ 0.88*
 \\
Error in test set nBYL after feedback
 & 65.48 $\pm$ 7.49
 & 16.16 $\pm$ 5.41*
 & 15.50 $\pm$ 7.23
 & 14.27 $\pm$ 5.87
 \\
\hline
\end{tabular}
\end{center}
\end{table*}

\section{Discussion}

\subsection{Fundamentals of the system}
We have presented the so-called AAN system that can replace an initial visually elicited reflex reaction by a feedback triggered action which uses concept-like entities. These concept like entities emerge through associative learning, where the first layer is receiving environmental feedback in an associative way and all other layers learn unsupervised. Visual inputs (presented as RGB images) are transformed into low-level visual features using a pre-trained convolutional neural network (named CNN above), which we use \textit{only} as a feature reservoir (in analogy to the early visual cortical processing areas in vertebrates).
Thus, the actual structure and learning procedure of the CNN are irrelevant for this study.

Unsupervised learning first makes sure to generate activity that gets sharper from layer to layer and is high in response to ``this'' and low in response to ``not this'' (related to \emph{closure}, layers 3). This activity is further processed by performing \emph{relational operations} ([Yellow \& Left] and [IF Yellow \& Left $\rightarrow$ Big]), finally triggering behavior via feedback. This feedback acts at the CNN and stimulates its neurons in a way as if a big entity had been visible. Thus, in some sense, the system ``imagines'' the Big entity. This together with the resulting virtual behavioral response represents a simple form of \emph{externalization}.

We used different learning rules to achieve this, all of which are common in real neural systems, but details of this are not central to our study. Hence, learning rules are only presented in the Appendix. Input-correlation-based (ICO) learning \cite{porr2006strongly} is used for the environmentally-controlled layers of the AAN system. In other layers Hebbian learning in some variants is used paired with an annealing procedure that limits weight growth. This mechanism is closely related to synaptic scaling \cite{turrigiano2008self,tetzlaff2011synaptic}.  These two different types of associative learning are used throughout the AAN system because this allows training the system using smaller data sets and is faster as compared to deep Reinforcement Learning and the here-obtained network structures are better interpretable than those obtained using deep learning. Importantly, we us a dendrite-like structure to allow for calculations combining different input signals. Also these types of structures are common in e.g., pyramidal cells, which can perform complex, local dendritic calculations \cite{payeur2019classes}.

\subsection{Evolutionary perspective}
It is arguable whether humans are the only living beings using concepts and the associated symbolic processing of reasoning (and communication). A possible indicator in pre-linguistic animals -- e.g., apes -- for employing symbolic processing could be whether or not they can develop and execute mental plans consisting of several action steps.  While planning for the future is a much agreed-upon fact in several species \cite{raby2007planning}, the knowledge about which mental processes drive planning is sparse. There is still a discussion to what degree seemingly deliberate decisions might indeed be just reflex driven. The system we have developed can help in formalizing the discussion of what one would call reflex- versus deliberation-driven. One could ask: If one observes a type of feedback that is based on relations between discrete entities, like in our system, is this a more advanced reflex, or is this an instance of primitive reasoning? Clearly here we are still at the level of ``implicit deliberation'' as classified by \cite{ledoux2018surviving} as such a system is not able to ``speak about''. Furthermore, in our system we do not yet achieve representations handling relations of other relations which is claimed by \cite{penn2008darwin} to be the main characteristic of human reasoning. However, we would argue that one could use systems as the one here as a stepping stone towards building more advanced systems and use those to analyze more complicated questions about human concept formation, too.

\subsection{Artificial system perspective}
Our system is developed in a grounded way, which is important when developing artificial agents that can act in a not predefined environment and adapt to it \cite{brooks1990elephants}. There are of course many grounded artificial developmental systems published \cite{cangelosi2010grounding,doya2019toward}, but all are domain-bound. Different from this, our system is essentially open. All mechanisms will work regardless of the type or modality of the inputs. We show how the system learns, using for concept development only the statistical structure existing in our artificial world. Hence, if there were more action-relevant intrinsic conjectures, a system with more neuronal pools could extract these without changing the general architecture.

The architecture of our system is block-wise pre-wired. Every block essentially represents a neuronal cell-assembly with certain (learned) properties. Pre-wiring between blocks omits the step of also having to learn the correct forward stream for the different modalities. Hence, pre-wiring shortcuts this and allows extracting Concepts rather quickly. The system will, thus, in the same way work efficiently in case of more complex conjectures, like (Yellow AND Left) AND (Triangle OR Square) if they exist in the data. Note, however, that we had shown in an older study that cell assemblies and their connectivity can also be developed by synaptic plasticity, which, however, requires prolonged self-organizing procedures \cite{tetzlaff2015use}.   Hence, doing this should work, but will be time consuming and would - in conjunction with the current study - not add to its core messages.

The development study \cite{ugur2015bottom} is designed to extract object categories in an exploratory way and to build
planning domains based on that. The first part, extraction of object categories, is comparable to our study, but it does not put emphasis on relation formation between categories and does not analyze distinction between categories and concepts. Furthermore, many aspects in their study are predefined and they use mostly non-biological mechanisms (of machine learning and data storage). In general, usage of complex predefined structures is prevalent in developmental robotics \cite{braud2020robot,gumbsch2019autonomous,bugur2019effect}, whereas our system has been designed from a bottom-up sensori perspective and by employing a stronger view onto biological realism with fewer predefined aspects.

\subsection{Limitations and Conclusion}

Our system uses only excitatory connections, where more complicated tasks may also require inhibitory connectivity. Furthermore, as mentioned above, our system is block-wise pre-wired where we had stated that developing a system with comparable functionality from randomly connected neurons would in principle be possible. By way of analogy: human development until arrival at more and more Concepts takes years, too, and a similar effect of prolonged learning times would be obtained if avoiding the short-cut of pre-wiring the system to some degree.

In addition, we are here only using classical approaches to neuronal systems, stopping short of the brain's dynamic complexity, but we show that even with this simple system we can develop models that learn primordial concepts and we think that this can serve as a valuable starting point for more complex neuronal approaches.

Our model world is (purposefully) quite simple. However, with the now-existing deep networks it is possible to pre-extract decisive feature representations (beyond those of our CNN layers) when confronted with more realistic input spaces coming from the real world. Doing this might allow investigating more complicated cognitive properties (e.g., considering relations of relations which is deemed to be important for human thought, as discussed above).

Finally, in our small neural network we do not consider making any intrinsic knowledge explicit via language. If possible, this route might lead eventually towards traits that point to artificial consciousness.

Hence future work could choose to address this and/or more complex ``worlds'', which, however, will be very demanding. While this may be a way forward, we would argue that we have presented here a neural-representation-based system that offers an model for Concept emergence within autopoietic systems that only receive information from the environment but not from external supervision processes.

\section{Appendix}
\subsection{Neural properties of the system}\label{section_neural}

We use the same activation equations for all neurons in our scheme (except CNN convolutional layers). First we calculate the weighted sum of the inputs:
\begin{equation}\label{eq_condition1}
y=\bm{\omega} ^T \mathbf{u},
\end{equation}
where $\mathbf{u}=(u_1,...,u_n)$ are inputs, $\bm{\omega}=(\omega_1,...,\omega_n)$ are weights,
$n=50$ and afterward apply sigmoidal saturation:
\begin{equation}\label{eq_sigfun}
v = f(y)=\frac{1}{1+e^{-a(y-b)}},
\end{equation}
where $a=0.1$, $b=100$.

Activations taken from the CNN can be specified as follows: The network structure, shown in Fig.~\ref{scheme}, was trained to distinguish 28 classes: 27 classes are 3 $\times$ 3 $\times$ 3 feature combinations (size, shape and color combinations) and one additional output is provided for left vs. non-left position of a shape. This output was added due to insufficient strength of the signal Left in the feature reservoir without this. For network training we used 27 $\times$ 900 images of each kind (as described above), which were independently generated for all sets (training or test sets) described above. However, those details are of secondary importance, as we only use this network in our architecture as a feature reservoir and a different network could be used here, too.

The network has the following convolutional layers (interspersed with four max pooling layers at the bottom of the network): RGB Image (100 $\times$ 100 $\times$ 3) $\rightarrow$ Layer 1 (96 $\times$ 96 $\times$ 4) $\rightarrow$ Layer 2 (44 $\times$ 44 $\times$ 32) $\rightarrow$ Layer 3 (18 $\times$ 18 $\times$ 64) $\rightarrow$ Layer 4 (5 $\times$ 5 $\times$ 128) $\rightarrow$ Layer 5 (3 $\times$ 3 $\times$ 128) $\rightarrow$ Layer 6 (1 $\times$ 1 $\times$ 64).

The AAN in our system is supplied with the activations of layers 4 to 6 (overall 4416 neurons). The convolutional layers have ReLU activation functions, which we normalize to the interval $[0,1]$ in our study, so that the scale is compatible with the sigmoidal neurons used in the other parts of our system.

Simulated motor activation is triggered in case the activity of all 300 layer 0 neurons in the Big 0 pool summed up exceeds threshold $F$, hence $\sum_{j=1}^{N} v_j > F$,  with $N=300$ and $F=20$. The same applies to Yellow 0 and Left 0.

\subsection{Learning Mechanisms}
In the AAN system we use four different learning mechanisms, which are all based only on the correlation between inputs. Hence, no explicit supervision is used in the AAN. This is meant to simulate ``simple'' brains, which have to rely exclusively on signals either from the environment (via their sensors) or arising intrinsically. We use:
\begin{enumerate}
        \item
    Input correlation learning (ICO) related to heterosynaptic plasticity \cite{porr2006strongly}
    \item
    Conventional Hebbian learning.
    \item
    Above average Hebbian learning related to mechanisms of synaptic scaling \cite{turrigiano2008self,tetzlaff2011synaptic}, where learning happens only for above average activity.
    \item
    Above average Hebbian/Anti-Hebbian balanced learning related to long-term potentiation and depression \cite{bear1994synaptic} also coupled to synaptic scaling \cite{turrigiano2008self,tetzlaff2011synaptic}.
\end{enumerate}

Some of these learning rules operate locally on a dendritic branch similar to local learning processes on dendrites in cortical pyramidal cells \cite{losonczy2008compartmentalized}.

Furthermore, we use the well-known physiological fact that weight growth in synapses is limited (large synapses grow less \cite{bi1998synaptic}) and implement this through a mechanism that reduces the learning rate when a synapse gets big (annealing of the learning rate).

Note that all here-used learning mechanisms are found in real neural system. An in-depth discussion of this would, however, exceed the scope of this paper.

We will now first describe the learning rules and afterward explain for which connections they are used and also summarize all layer-specific parameters in the related tables.
\subsubsection{Input correlation learning (ICO)}\label{section_ICO}



\begin{table}[ht]
\begin{center}
\caption{Parameters for ICO and Hebbian learning.}
\label{combi-tab}
\begin{tabular}{|l|l|l|l||l|l|ll|}
\hline
\multirow{2}{*}{ICO} & \multirow{2}{*}{$\mu^{i_p}$} & \multirow{2}{*}{$\mu^{i_n}$} & \multirow{2}{*}{$\Phi$} & \multirow{2}{*}{Hebb} & \multirow{2}{*}{$\mu(0)$} & \multicolumn{2}{l|}{Annealing}   \\ \cline{7-8}
                     &                       &                       &                      &                       &                        & \multicolumn{1}{l|}{$\Theta$} & $c$   \\ \hline
                     & 80                    & 34                    & 0.01                 &                       & 1.0                    & \multicolumn{1}{l|}{0.99}  & 0.9 \\ \hline
\end{tabular}
\end{center}
\end{table}

This type of learning relies on the correlation between the to-be-learned input correlated with a pre-wired reflex-inducing input. For example, a big (dangerous) shape triggers a reflex (of running away) by looming over an agent with a bit of a delay relative to the moment the image appears. ICO learning \cite{porr2006strongly} makes use of the correlation between these two signals (image input presentation and the looming signal), where a positive correlation leads to weight growth and vice versa. We use a slightly modified version of ICO with:
\begin{equation}
  \bm{\omega} \leftarrow
    \begin{cases}
     \textrm{ $\bm{\omega} + \mu^{i_p} \mathbf{u^{\Phi}}$} & \text{Reaction required, but not present}\\

       \textrm{ $\bm{\omega} - \mu^{i_n} \mathbf{u}^{\Phi}$} & \text{Reaction present, but not required},
    \end{cases}
\end{equation}
for the synaptic weight vector $\bm\omega$, with $\bm{\omega}(0)=0$ and input-determined vector $\mathbf{u^{\Phi}}=(u^{\Phi}_1,\dots,u^{\Phi}_n)$, where:
\begin{equation}
  u^{\Phi}_i =
    \begin{cases}
     \textrm{ $u_i$} & \textrm{if $u_i>\Phi$}\\
      0 & \text{otherwise}.
    \end{cases}
\end{equation}

Parameters used for ICO rule are provided in Table~\ref{combi-tab},~left. Note, that the ICO rule is intrinsically convergent, because learning will stop as soon as the learned synapse is strong enough to supersede the reflex. For proof of this property see \cite{porr2003isotropic}. This rule is a typical heterosynaptic plasticity rule, which are common e.g., in the Hippocampus \cite{oh2015heterosynaptic}.

\subsubsection{Conventional Hebbian learning}\label{section_hebbian}


For some of the connections we use regular Hebbian learning given by:
\begin{equation}\label{eq_hebbian}
\bm{\omega} \leftarrow \bm{\omega} + \bm{\mu^{h}(t)} \mathbf{u}v,~with~\bm{\omega}(0)=0
\end{equation}

Stability of learning is assured by an annealing process of the learning rate in case the neuron produces very high output. This annealing process leads to the final stopping of learning. It is related to the physiological property of limited weight growth, where large synapses do not continue to grow any longer \cite{bi1998synaptic}. We define:
\begin{equation}\label{eq_mudecay}
\text{if } v_j(t)>\Theta\text{,}~\mu^{h}_j(t)=c\mu^{h}_j(t-1),
\end{equation}
with $j=1,...,N$, where $N$ is the number neurons in a pool, $\Theta$ is the threshold and $c$ defines the rate of annealing ($c<1$). Note that we define $t=1,2,...,t_{max}$ as the index for the discrete sequence of image presentations. Parameter values are provided in Table~\ref{combi-tab},~right.

\begin{table*}[ht]
\begin{center}
\caption{Parameters for above average Hebbian learning. Numbers 1,2,3 refer to the layer indices. Note, that layer 1 needs slower annealing, as its inputs are wider distributed (see histograms of layer 0 in Fig.~\ref{histos}) than in the other layers.}
\label{table_param_hebb_aa}
\begin{tabular}{|l|l|l|l|c|c|c|c|c|c|c|c|}
\hline
\multirow{2}{*}{\begin{tabular}[c]{@{}l@{}}AA\\ Hebb\end{tabular}} & \multirow{2}{*}{\begin{tabular}[c]{@{}l@{}}$\mu(0)$\\ 1,2,3\end{tabular}} & \multirow{2}{*}{\begin{tabular}[c]{@{}l@{}}$\mu(0)$\\ AND 1\end{tabular}} & \multirow{2}{*}{\begin{tabular}[c]{@{}l@{}}$\mu(0)$\\ AND 2\end{tabular}} & \multicolumn{2}{c|}{$U$} & \multicolumn{2}{c|}{$V$} & \multicolumn{2}{l|}{Annealing, 1} & \multicolumn{2}{l|}{Annealing, rest} \\ \cline{5-12}
&                                                                     &                                                                    &                                                                    & $g_U$         & $r_U$        & $g_V$         & $r_V$        & $\Theta$            & $c$              & $\Theta$              & $c$               \\ \hline
\multicolumn{1}{|c|}{}                                             & \multicolumn{1}{c|}{5}                                              & \multicolumn{1}{c|}{0.05}                                          & \multicolumn{1}{c|}{0.025}                                         & 0.9        & 1.0       & 0.9        & 0.1       & 0.99             & 0.9            & 0.995              & 0.1             \\ \hline

\end{tabular}
\end{center}
\end{table*}

\begin{table}[ht]
\begin{center}
\caption{Parameters for above average Hebbian/Anti-Hebbian balanced learning.}
\label{table_params_balanced_hebb}
\begin{tabular}{|l|c|c|c|c|c|c|c|}
\hline
\multirow{2}{*}{\begin{tabular}[c]{@{}l@{}}Balanced \\ Hebb\end{tabular}} & \multirow{2}{*}{$\mu(0)$} & \multicolumn{2}{c|}{$U$} & \multicolumn{2}{c|}{$V$} & \multicolumn{2}{c|}{Annealing} \\ \cline{3-8}
                                                                          &                     & $g_U$          & $r_U$         & $g_V$          & $r_V$         & $\Theta$           & $c$            \\ \hline
                                                                          & 0.05                & 0.9        & 1.0       & 0.9        & 1.0       & 0.995           & 0.1          \\ \hline
\end{tabular}
\end{center}
\end{table}

\subsubsection{Above average Hebbian learning}\label{section_Input}

In this type of learning weights will be strengthened only if inputs and outputs exceed average past activation. The use of average past activation is related to mechanisms of \textit{synaptic scaling}, where weight growth is determined by the activation level of the neuron \cite{turrigiano2008self,tetzlaff2011synaptic}. In our case this allows bootstrapping, because we start with near zero weight and, hence, activations are also initially near zero. Thus, without this type of synaptic scaling, learning would not start.

We define:
\begin{equation}\label{eq_inputonly}
\bm{\omega} \leftarrow \bm{\omega} + \bm{\mu^{aa}(t)} \mathbf{u^{\Delta}}H(v^{\Delta}),~with~\bm{\omega(0)}=0.01,
\end{equation}
where $H$ is the Heaviside function and $\mathbf{u^{\Delta}}=(u^{\Delta}_1,\dots,u^{\Delta}_n)$, where:
\begin{equation}
  u^{\Delta}_i =
    \begin{cases}
     \textrm{ $u_i-r_U U_i(t)$} & \textrm{if $u_i>r_U U_i(t)$}\\
      0 & \text{otherwise}
    \end{cases}
\end{equation}
The function $U$ represents the average past activation and is defined by: $U_i(0)=0$ and $U_i(t)=g_U U_i(t-1)+(1-g_U)u_i(t)$ which calculates a sliding average, with $g_U$ the weighing factor of the averaging history and $r_U$ the amplitude of the influence of $U$. The function $v^{\Delta}$ is defined in the same way as $u^{\Delta}_i$.

Hence, learning only takes place if inputs and outputs are strong enough. In addition and different from conventional Hebbian learning, we use the Heaviside function (Eq.~\ref{eq_inputonly}) and not the actual value of $v$ for correlation with the input. Hence, learning strength is driven directly by the value of the input $u$.

Also for this rule we use annealing of the learning rate, defined as in Eq.~\ref{eq_mudecay} above. All parameter values are given in Table~\ref{table_param_hebb_aa}.

\subsubsection{Above average Hebbian/Anti-Hebbian balanced learning} \label{section_LTD}

For this type of learning, we implement weight reduction, related to long-term depression (LTD) in real neurons, to reduce the weights in those inputs which do not support the operation desired by the considered neuron, whereas the weights of the other synapses will grow (long-term potentiation, LTP, \cite{bear1994synaptic}):
\begin{equation}
\label{eq_outputsign}
\bm{\omega} \leftarrow \bm{\omega} + \bm{\mu^{b}(t)} \mathbf{u^{\Delta}}sign(v^{\Delta}),~with~\bm{\omega(0)}=0.01
\end{equation}

Terms $\bm{u^{\Delta}}$ and $v^{\Delta}$ are defined as above. The $sign$ function takes a similar role as the Heaviside function above, but now weights can grow or shrink dependent on the output.

Also for this rule we use annealing of the learning rate, defined as in Eq.~\ref{eq_mudecay} above. Parameter values are given in Table~\ref{table_params_balanced_hebb}.

\subsection{Application of the different learning rules and parameters}

The different learning rules have to be used at different target layers to assure correct system behavior.

\underline{Feed-forward paths:} Input correlation learning (ICO) is used for all connections that converge on layer 0 neurons (see Fig.~\ref{scheme}). This leads to the first step of separating features from each other but the resulting distributions are still rather dispersed (Fig.~\ref{histos}).

For all other feed-forward paths (blue arrows in Fig.~\ref{scheme}) we use Above Average Hebbian Learning. This, in combination with the stopping mechanism, leads to a substantial sharpening of the distributions due to self-organization which leads to a kind of soft winner-takes-all mechanisms favoring all stronger signals.

This type of learning also happens at the feed-forward branch of the dendrite of Big 3 (left branch in Fig.~\ref{scheme}).

\underline{Feedback paths:} The right branch in Big 3 uses balanced (Hebbian/Anti-Hebbian) learning (Fig.~\ref{scheme}). The utility of this type of learning here is that only neurons obeying IF-THEN rule have possibility to enhance connections to the branch. Without that the weights would fall to zero.

The final feedback onto the convolutional layers (Fig.~\ref{scheme}) uses conventional Hebbian learning, where annealing (Eq.~\ref{eq_mudecay}) depends on the maximum output in the pool.

\underline{Local learning at dual dendrites:} This case concerns the convergence of feed-forward with feedback signals at the convolutional layers 4-6 and in the pool Big 3. In general, dendritic branches learn \textit{independently} by rules, which are different for the feed-forward and the feedback branches as described above. This is required to assure that both input-types can exert an influence on the output. If input-types were pooled for learning, the weaker input would never grow. Note that the feed-forward dendritic branch of the convolutional layers receives  activations, which are pre-trained and do not change in the course of learning of the AAN network. Local learning on different branches is a common phenomenon found at the dendrites of cortical pyramidal cells \cite{cichon2015branch}.

\underline{Input integration processes at dual dendrites:} While learning rules are different at the branches, they still use the same output $v$ for driving their learning rules obtained by using the maximum operation of the saturated sums $v_1=f(\bm{\omega_1} ^T \mathbf{u_1})$ and $v_2=f(\bm{\omega_2} ^T \mathbf{u_2})$ for the two branches,  $v=max(v_1,v_2)$.   The maximum operator at the junction of the branches is related to well-know gating mechanisms that happen also at dendritic structures \cite{payeur2019classes}.
\end{CJK}
\bibliographystyle{ieeetr}
\bibliography{Tamosiunaite_et_al_w_supplement}

\newpage

\section{Supplement}

In this supplementary material we are providing additional results. In Section~1 we show statistics on network activation. In Section~2 we quantify learning.

\begin{table*}[ht]
	\begin{center}
		\caption{Discretization of signalling in layers 1-3. Percentage of low neuronal activation $(<0.1)$ (averaged over 300 neurons each) in case of images not belonging to category Big, Yellow, or Left (true negatives) and of high activation $(>0.9)$ in case of images belonging to the category (true positives). Percentage and standard deviation over 20 trials are shown. Asterisks denote significant increase in value from previous level to the level with the asterisk ($p<0.05$ for the t-test, 20 repetitions). TS denotes test set.}
		\label{tab_sharpening}
				\begin{tabular}{lllll}

					\multirow{1}{*}{} & \multicolumn{1}{l}{Condition
					} & \multicolumn{1}{l}{Layer 1
					} & \multicolumn{1}{l}{Layer 2
					} & \multicolumn{1}{l}{Layer 3
					} \\\cline{2-5}
					\hline
					\multirow{3}{*}{Big} & \multicolumn{1}{l}{Baseline TS: False $<$ 0.1
					} & \multicolumn{1}{l}{90.91 $\pm$ 4.55
					} & \multicolumn{1}{l}{96.61 $\pm$ 2.94 *
					} & \multicolumn{1}{l}{97.82 $\pm$ 2.25
					} \\\cline{2-5}
					& \multicolumn{1}{l}{Baseline TS: True $>$ 0.9
					} & \multicolumn{1}{l}{88.62 $\pm$ 3.36
					} & \multicolumn{1}{l}{99.23 $\pm$ 0.62 *
					} & \multicolumn{1}{l}{99.72 $\pm$ 0.21 *
					}\\\cline{2-5}
					& \multicolumn{1}{l}{nBYL TS: False $<$ 0.1
					} & \multicolumn{1}{l}{70.99 $\pm$ 5.38
					}& \multicolumn{1}{l}{78.87 $\pm$ 4.54 *
					} & \multicolumn{1}{l}{81.41 $\pm$ 4.42
					} \\\cline{1-5}
					
					\hline
					\multirow{3}{*}{Yellow} & \multicolumn{1}{l}{Baseline TS: False $<$ 0.1} & \multicolumn{1}{l}{67.32 $\pm$ 10.03
					} & \multicolumn{1}{l}{87.37 $\pm$ 8.90 *
					} & \multicolumn{1}{l}{91.65 $\pm$ 7.79
					} \\\cline{2-5}
					& \multicolumn{1}{l}{Baseline TS: True $>$ 0.9} & \multicolumn{1}{l}{79.36 $\pm$ 8.41
					} & \multicolumn{1}{l}{99.47 $\pm$ 1.03 *
					} & \multicolumn{1}{l}{99.91 $\pm$ 0.39
					}\\\cline{2-5}
					& \multicolumn{1}{l}{nBYL TS: True $>$ 0.9} & \multicolumn{1}{l}{29.53 $\pm$ 14.83
					}& \multicolumn{1}{l}{74.23 $\pm$ 12.72 *
					} & \multicolumn{1}{l}{86.18 $\pm$ 9.01 *
					} \\\cline{1-5}
					
					\hline
					\multirow{3}{*}{Left} & \multicolumn{1}{l}{Baseline TS: False $<$ 0.1} & \multicolumn{1}{l}{92.49 $\pm$ 6.61
					} & \multicolumn{1}{l}{97.18 $\pm$ 3.73 *
					} & \multicolumn{1}{l}{98.01 $\pm$ 2.83
					} \\\cline{2-5}
					& \multicolumn{1}{l}{Baseline TS: True $>$ 0.9} & \multicolumn{1}{l}{97.44 $\pm$ 1.62
					} & \multicolumn{1}{l}{99.37 $\pm$ 0.37 *
					} & \multicolumn{1}{l}{99.51 $\pm$ 0.25
					}\\\cline{2-5}
					& \multicolumn{1}{l}{nBYL TS: True $>$ 0.9} & \multicolumn{1}{l}{81.17 $\pm$ 0.85
					}& \multicolumn{1}{l}{81.85 $\pm$ 7.60
					} & \multicolumn{1}{l}{82.53 $\pm$ 7.46
					} \\\cline{1-5}

				\end{tabular}
		\end{center}
	\end{table*}
	
	\subsection{Statistics on network activation}
	
	In Table~\ref{tab_sharpening} we present statistics quantifying the discretization effect seen in Figure 4 of the main text. We show which percentage of images that are \emph{not belonging} to a category (Big, Yellow or Left) has a low neuronal response $(<0.1)$  and which percentage of images \emph{belonging} to a category  has a high neuronal response $(>0.9)$. One can see that those percentages consistently grow from layer 1 to layer 3, however the change from layer 2 to 3 is small and most of the times not statistically significant.
	
	\begin{table*}[ht]
		\begin{center}
			\caption{Performance of relational operations on the feedback path. Percentage of low neuronal activation $(<0.1)$ (averaged over 300 neurons each) in case of images not belonging to category (true negatives) and of high activation $(>0.9)$ in case of images belonging to the category (true positives). ``AND 1'' denotes Yellow \& Left 1 in the first three rows, but Magenta \& Left 1 in the last three rows.  The same for the ``AND 2''.  Percentage and standard deviation over 20 trials are shown. TS denotes test set.}
			\label{tab_ifthen}
					\begin{tabular}{lllll}

						\multirow{1}{*}{} & \multicolumn{1}{l}{Condition
						} & \multicolumn{1}{l}{Layer ``AND 1''
						} & \multicolumn{1}{l}{Layer ``AND 2''
						} & \multicolumn{1}{l}{Branch 2 Big 3
						} \\\cline{2-5}
						\hline
						\multirow{3}{*}{Yellow \& Left} & \multicolumn{1}{l}{Baseline TS : False $<$ 0.1
						} & \multicolumn{1}{l}{95.43 $\pm$ 6.65
						} & \multicolumn{1}{l}{99.91 $\pm$ 0.00
						} & \multicolumn{1}{l}{99.91 $\pm$ 0.00
						} \\\cline{2-5}
						& \multicolumn{1}{l}{Baseline TS: True $>$ 0.9
						} & \multicolumn{1}{l}{99.14 $\pm$ 0.08
						} & \multicolumn{1}{l}{100.00 $\pm$ 0.00
						} & \multicolumn{1}{l}{\textbf{100.00 $\pm$ 0.00}
						}\\\cline{2-5}
						& \multicolumn{1}{l}{nBYL TS: True $>$ 0.9
						} & \multicolumn{1}{l}{84.13 $\pm$ 8.55
						}& \multicolumn{1}{l}{84.66 $\pm$ 8.43
						} & \multicolumn{1}{l}{84.66 $\pm$ 8.43
						} \\\cline{1-5}
						
						\hline
						\multirow{3}{*}{Magenta \& Left} & \multicolumn{1}{l}{Baseline TS: False $<$ 0.1 } & \multicolumn{1}{l}{98.06 $\pm$ 4.17
						} & \multicolumn{1}{l}{99.68 $\pm$ 0.30
						} & \multicolumn{1}{l}{99.96 $\pm$ 0.01
						} \\\cline{2-5}
						& \multicolumn{1}{l}{Baseline TS: True $>$ 0.9 } & \multicolumn{1}{l}{99.12 $\pm$ 0.32
						} & \multicolumn{1}{l}{99.12 $\pm$ 0.32
						} & \multicolumn{1}{l}{\textbf{9.91 $\pm$ 29.74}
						}\\\cline{2-5}
						& \multicolumn{1}{l}{nBYL TS: False $<$ 0.1 } & \multicolumn{1}{l}{4.53 $\pm$ 1.84
						}& \multicolumn{1}{l}{83.78 $\pm$ 32.9
						} & \multicolumn{1}{l}{99.98 $\pm$ 0.01
						} \\\cline{1-5}

					\end{tabular}
			\end{center}
		\end{table*}
		
		In Table~\ref{tab_ifthen} we present statistics quantifying the relational calculations for the blocks positioned on the right in Figures 3 and 4 of the main text (Yellow \& Left 1, Yellow \& Left 2 and Branch 2 Big 3). We show which percentage of images that are \emph{not belonging} to a category (e.g. Yellow \& Left) has a low neuronal response $(<0.1)$  and which percentage of images \emph{belonging} to that category  has a high neuronal response $(>0.9)$. In the top three rows of the table we show results for the original setting, while in the bottom three rows we show results for the control setting, where instead of the yellow stream the magenta stream is used. As stated in the main text, there is a clear asymmetry in the world of an agent that yellow figures on the left are most of the time big. However for magenta figures on the left there is no well  expressed asymmetry in respect to size. Thus, in the control architecture the feedback loop shall not get activated. However, let us first analyze layers ``AND 1'' and ``AND 2'', which shall not differ between the two architectures. Values in the first two columns are high (except row six, where high values are achieved only after sharpening in the column ``AND 2''). This shows that the conjecture ``AND'' is detected with high accuracy when it exists and rejected with high accuracy in case such conjecture does not exist. Main difference emerges in column 3 labelled ``Branch 2 of the neuron Big 3''. Here the ``IF Yellow \& Left THEN Big'' conjecture emerges with 100 \% accuracy (see bold font value in the last column, second row of the table). Similar holds for the nBYL test set, where the value is slightly smaller due to the reasons described in the main text and reaches 84\%. However in case of Magenta \& Left the conjecture ``IF Magenta \& Left THEN Big'' is not strongly expressed (beneath 10\%, see bold font in line 5). This value can be brought even smaller with slightly tuning the threshold $r_V$ for the above average Hebbian/Antihebbian balanced learning. Throughout this work $r_V=1.0$ was used without any tuning of this parameter. Note in addition, that in the test set nBYL all images are non-big yellows on the left. Thus for the original architecture (top three rows) only the positive condition for the nBYL test set is measured. However, all images from nBYL obey the conjecture \textit{not} Magenta \& Left. Due to that, for the control architecture in the bottom three rows the negative condition is measured. 
		
		\subsection{AAN learning aspects}
		
		The AAN network has two learning modes: environmentally supervised learning in layer 0 and unsupervised learning in the remaining parts of the network. We show how learning progresses, separately for the environmentally supervised (ICO) and for the unsupervised learning in the subsections below.
		
		\subsubsection{Quantification of environmentally supervised learning}
		
		\begin{figure*}[ht]
			\centering \includegraphics[width=15cm]{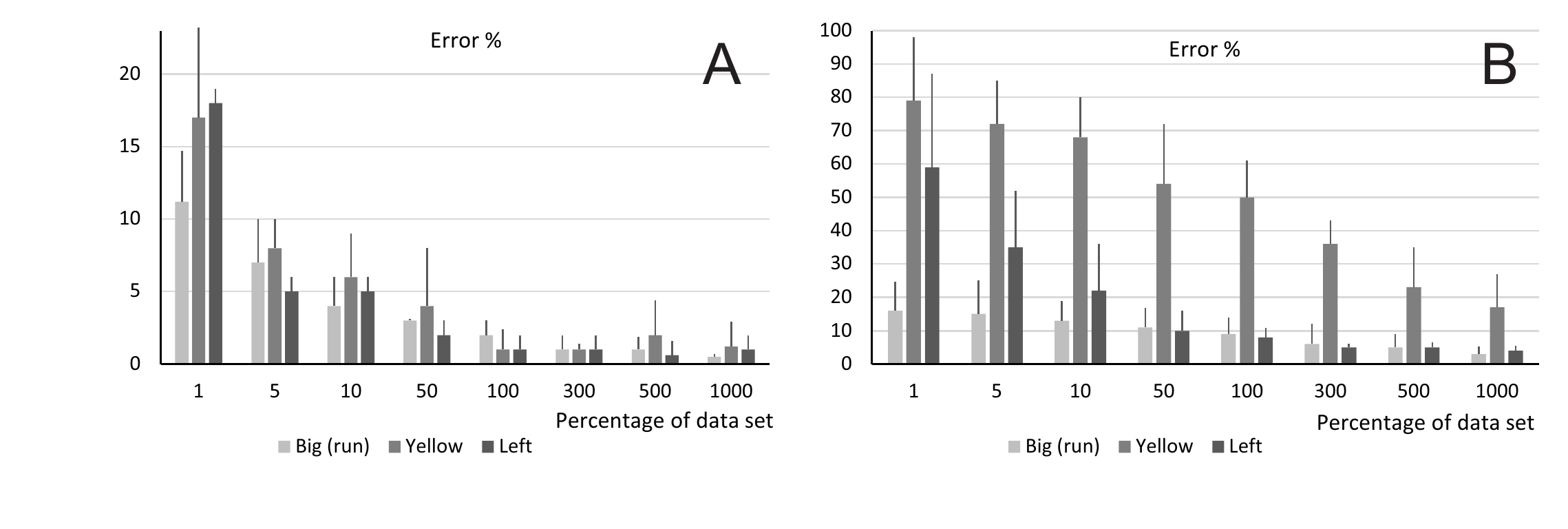}
			\caption{Progress in ICO learning. Tested on data set: A) \textbf{Baseline}, B) \textbf{nBYL}. On the horizontal axis, percentage of data set larger than 100 means repeated presentations of the training data set, each time in different randomized order.}
			\label{learning_ICO}
		\end{figure*}
		
		The layer 0 is trained using ICO learning as described in the Appendix of the main paper.  The errors for three entities: Big (run), Yellow and Left are provided in Fig.~\ref{learning_ICO}. One can see that for the data set \textbf{baseline} (Fig.~\ref{learning_ICO}~A) learning effects are already substantial after just a couple of percents of the training data set has been presented. The error stabilizes after presenting the training data set once (100 percent on the graph). Note that ICO learning reacts to errors and, thus, weights change as soon as an error is encountered. Thus, changes happen frequently at the beginning, but less and less frequently as learning progresses. In idealized condition one would achieve error-free performance and complete stopping of learning after some time \cite{porr2006strongly}. In our more complicated task, similar to the real world, the system stabilizes at the level, where errors happen infrequently and learning continues to happen at a low rate life-long. For the test set \textbf{nBYL}, learning is slower (see Fig.~\ref{learning_ICO}~B), as there are very few images of this type in the training set. Especially slow is the learning for Yellow when based on the \textbf{nBYL} test set. This was shown to be the limiting factor for the network to correctly judge the rule [IF Yellow \& Left THEN Big], in the main text in Figure 6.
		
		
		\subsubsection{Quantification of learning progress of unsupervised learning}
		
		\begin{figure}[ht]
			\begin{center}
				\includegraphics[width=9cm]{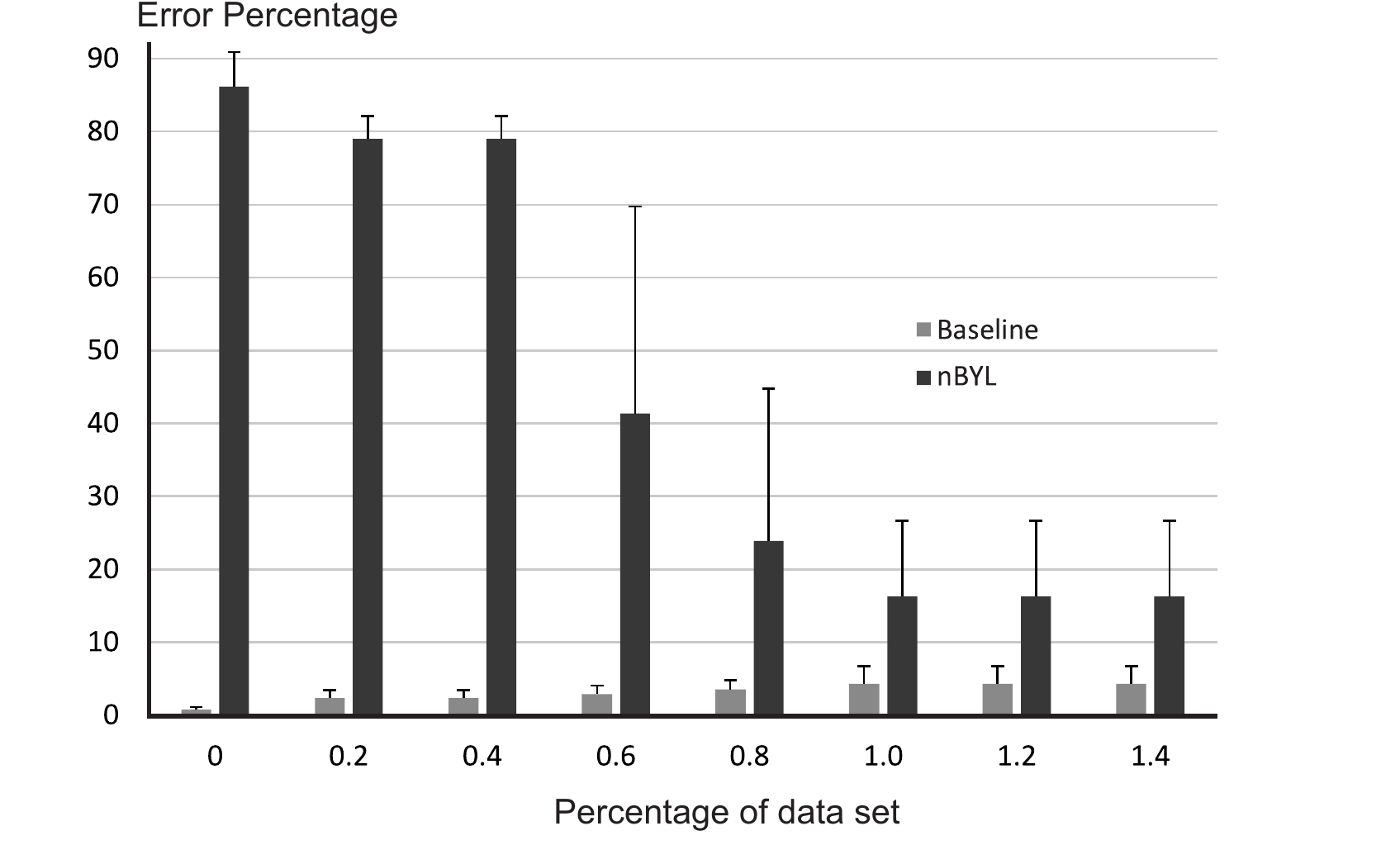}
				\caption{Progress of learning success. Percentage of data set larger than 100 means repeated presentation of the training data set, in different randomized order. Parameter values as given in the Appendix of the main text (Tables IV-VII) are used.}
			\end{center}
			\label{learning_prog_suppl}
		\end{figure}
		
		Here we quantify learning progress of the unsupervised components, based on the final outcome: percentage of correct running events tested using \textbf{baseline} and \textbf{nBYL} test sets (see Fig.~2). There we first show the error percentages evaluated on the two test sets after ICO learning, but before learning of the unsupervised components (percentage 0). Further we show the error percentages when stopping unsupervised learning early, after presenting training samples partially (percentages 20-80\%) or completely (100\%) or with partial repetitions (percentages $>$100\%).  For the \textbf{baseline} test set the errors are minimal after the ICO learning in layer 0, while for the \textbf{nBYL} test set errors are initially high. These errors reach their minimum (plateaus) at 100\%, which denotes that the full training set had been used ($=2742$ images). Training with repeating those images (above 100\%) does not anymore improve performance. Adding different images could potentially lead to still smaller errors, but optimizing the system along these lines is not in the core of  this study.
		
		To guarantee convergence of learning in case of parameter variations (e.g. slower learning rate, slower annealing) all results provided in the main text are obtained using longer learning sessions as compared to what is provided in  Fig.~2. For that 13710 iterations (500 percent, i.e. 5 times the learning set) was used.

\end{document}